\documentclass[10pt,journal,compsoc,twoside]{IEEEtran}

\usepackage{graphicx} \graphicspath{{img/}}
\usepackage[nocompress]{cite}
\usepackage{amsmath,amsfonts}
\usepackage{microtype}
\usepackage{booktabs}
\usepackage{balance}
\usepackage{dblfloatfix}

\usepackage{array}

\usepackage[pdftex, bookmarksnumbered=true, pagebackref=false]{hyperref}
\hypersetup{colorlinks,
citecolor=[rgb]{0.0,0.0,0.4},
filecolor=[rgb]{0.6,0.0,0.4},
linkcolor=[rgb]{0.0,0.0,0.4},
urlcolor=[rgb]{0.0,0.0,0.4}}
\makeatletter
\newcommand{\printfnsymbol}[1]{%
  \textsuperscript{\@fnsymbol{#1}}%
}
\makeatother

\begin{document}

\title{Going Deeper Into Face Detection: A Survey}

\author{Shervin~Minaee,
        Ping~Luo, 
        Zhe~Lin, 
        Kevin~Bowyer
\IEEEcompsocitemizethanks{\IEEEcompsocthanksitem Shervin~Minaee is a Machine Learning Lead at Snap Inc.
\IEEEcompsocthanksitem Ping~Luo is an assistant professor at the University of Hong Kong.
\IEEEcompsocthanksitem Zhe~Lin is a senior principal scientist at Adobe Research.
\IEEEcompsocthanksitem Kevin~Bowyer is the Schubmehl-Prein Family Professor of Computer Science and Engineering at the University of Notre Dame.}
}


\IEEEtitleabstractindextext{%
\begin{abstract}
Face detection is a crucial first step in many
facial recognition and face analysis systems. 
Early approaches for face detection were mainly based on classifiers built on top of hand-crafted features extracted from local image regions, such as 
Haar Cascades and Histogram of Oriented Gradients.  
However, these approaches were not powerful enough to achieve a high  accuracy on images of from uncontrolled environments.
With the breakthrough work in image classification using deep neural networks in 2012, there has been a huge paradigm shift in face detection. Inspired by the rapid progress of deep learning in computer vision, many deep learning based frameworks have been proposed for face detection over the past few years, achieving significant
improvements in accuracy.
In this work, we provide a detailed overview of some of the most representative deep learning based face detection methods by grouping them into a few major categories, and present their core architectural designs and accuracies on popular benchmarks.
We also describe some of the most popular face detection datasets.
Finally, we discuss some current challenges in the field, and suggest potential future research directions.
\end{abstract}
}
\maketitle

\IEEEdisplaynontitleabstractindextext
\IEEEraisesectionheading{\section{Introduction}
\label{sec:introduction}}

Face detection is an essential early step for 
tasks such as face recognition, facial attribute classification, face editing, and face tracking, and its performance has a direct impact on the effectiveness of those tasks \cite{facerec_survey, minaee2019biometric}.
Although great improvements have been made in uncontrolled face detection over the past few decades, accurate and efficient face detection in the wild remains an open challenge.
This is due to factors such as variations in poses, facial expressions, scale, illumination, image distortion, face occlusion, and other factors. 
Different from generic object detection, face detection features smaller variations in the aspect ratio, but much larger variations in scale (from several pixels to thousand pixels).

\begin{figure*}[h]
\centering
\includegraphics[width=0.8\linewidth]{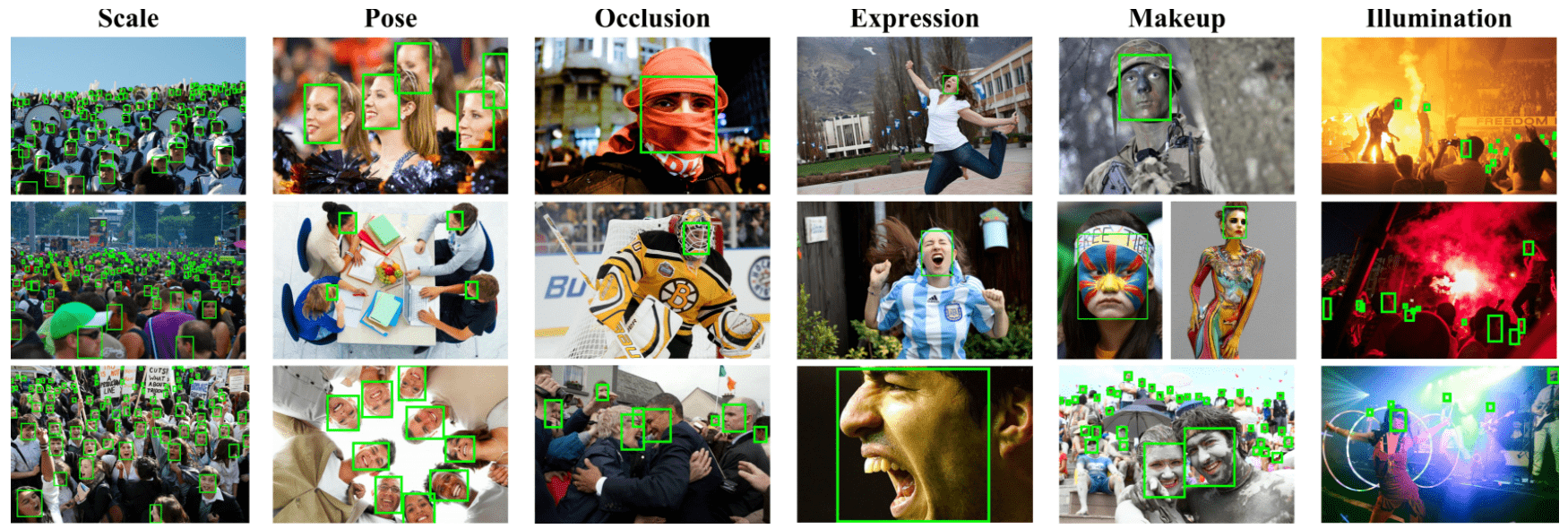}
\caption{Sample images from the ``Wider-Face'' face detection datasets, showing different variation factors. Courtesy of \cite{yang2016wider}}
\label{fig:wider_intro}
\end{figure*}

Early face detection efforts were mainly based on the classical approach, in which hand-crafted features were extracted from the image (or from  sliding windows on the image) and were fed into a classifier (or ensemble of classifiers) to detect  likely face regions. 
Two landmark classical works for face detection are the Haar Cascades classifier \cite{viola2001rapid} and the Histogram of Oriented Gradients (HOG) followed by SVM \cite{dalal2005histograms}.
These works represent great improvements on the state-of-the-art at their time.
However, face detection accuracy was still limited
on challenging images 
with multiple variation factors such as the ones shown in Fig \ref{fig:wider_intro}.

With the great success of deep learning in computer vision, researchers have
proposed
several promising model architectures 
over the past 6-7 years. 
Inspired by the cascade of classifiers idea, many of the earlier deep-learning-based models were based on Cascade-CNN architectures. 
But with the introduction of several novel architectures for general object detection, many more recent deep-learning-based models have shifted toward single-Shot Detection, R-CNN based architectures, feature pyramid network (FPN) models, and beyond.

Major surveys of face detection research up to circa 2000 include those by Yang et al. \cite{Yang_2002}, Rowley et al.s \cite{Rowley_1998}, and Hjelmås and Low \cite{Hjelmas_2002}.
Zhang and Zhang survey progress in face detection over roughly the next decade, to about 2010 \cite{Zhang_2010}.
Zafeiriou et al \cite{Zafeiriou_2015} survey face detection research over roughly the next five years, to near the beginning of the deep learning wave, around 2015.  
One of their conclusions is that “even when allowing a relatively large number of false positives (around 1,000), there are still around 15-20\% of faces that are not detected.’’
Our survey picks up where \cite{Zafeiriou_2015} ends, and covers the rapid progress in face detection from the beginning of the deep learning wave through the current time.
Table \ref{tab:summary_survey} summarizes and compares the existing surveys with our work.

\begin{table*}[t]
\caption{Comparisons of face detection surveys since 1998.}
\label{tab:summary_survey}
\begin{center}
\begin{tabular}{  m{0.5cm} | m{4cm}| m{3cm} | m{1cm} | m{1cm}|m{6cm}} 
\hline
No. & Survey Title & Reference & Year & Venue & Main Content \\ 
\hline
\hline
1 & Neural network-based face detection & Rowley et al. \cite{Rowley_1998} & 1998 & PAMI & A survey of 2-layered neural network for face detection.\\
\hline
2 & Face detection: A survey & Hjelmås and Low \cite{Hjelmas_2002} & 2001 & CVIU & A survey of traditional feature-based methods such as low-level cues (\emph{e.g.} edges) and active shape models (\emph{e.g.} Snakes), as well as image-based methods such as linear subspace methods. \\
\hline
3 & Detecting  faces in images: a survey & Yang et al. \cite{Yang_2002} & 2002 & PAMI & A survey of face detection from
a single image, focusing on feature engineering and  conventional classifiers such as EigenFace, Naive Bayes, and Support Vector Machine.\\
\hline
4 & A  survey  of  recent  advances  in  face detection & Zhang et al. \cite{Zhang_2010} & 2010 & Technical Report & A survey of Viola-Jones face detector and its variants.\\
\hline
5 & A survey on face detection in the wild: past, present and future & Zafeiriou et al. \cite{Zafeiriou_2015}& 2015 & CVIU & A survey of handcrafted features, boosting and Support Vector Machine, deformable models in face detection before the wave of deep learning.\\
\hline
6 & Going deeper into face detection: a survey (this work) & -- & 2021 & -- & A survey  of  the  recent  advanced deep-learning-based face detection, including more than fifty deep models for face detection.\\
\hline
\end{tabular}
\end{center}
\end{table*}

This paper provides a survey of the recent literature in deep-learning-based face detection, including more than fifty such detection methods.
It provides a comprehensive review with insights into different aspects of these methods, including the training data, choice of network architectures, loss functions, training strategies, and their key contributions. 
These works are organized into the following categories, based on their main technical contributions to face detection:
\begin{enumerate}
    \item Cascade-CNN Based Models
    \item R-CNN and Faster-RCNN Based Models
    \item Single Shot Detector Models
    \item Feature Pyramid Network Based Models
    \item Other models
\end{enumerate}

The remainder of this survey is organized as follows:
Section~\ref{sec:DL_model_overview} overviews popular Deep Neural
Network (DNN) architectures that serve as the backbones of many
modern face detection algorithms.
Section~\ref{sec:DL_models} reviews the
most significant state-of-the-art deep learning based face detection models, and their main technical contributions.
Section~\ref{sec:datasets} summarizes the most popular benchmarks for face detection, their size and other characteristics. 
Section~\ref{sec:performance} lists popular metrics for evaluating deep-learning-based face detection models and also tabulates the performance of models on those datasets.
Section~\ref{sec:challenges} discusses the main challenges and opportunities of deep learning-based face detection.
Section~\ref{sec:conclusions} presents our conclusions.

\section{Overview of Popular Deep Learning Architectures}
\label{sec:DL_model_overview}
This section provides an overview of prominent DNN architectures used by the computer vision community, including convolutional neural networks, recurrent neural networks and long short-term memory, encoder-decoder and autoencoder models, and generative adversarial networks. Due to space limitations, several other DNN architectures that have been proposed,
such as transformers, capsule networks, gated recurrent units, and
spatial transformer networks, 
are not covered.

\subsection{Convolutional Neural Networks (CNNs)}
\begin{figure}[h]
\centering
\includegraphics[page=3,width=0.98\linewidth]{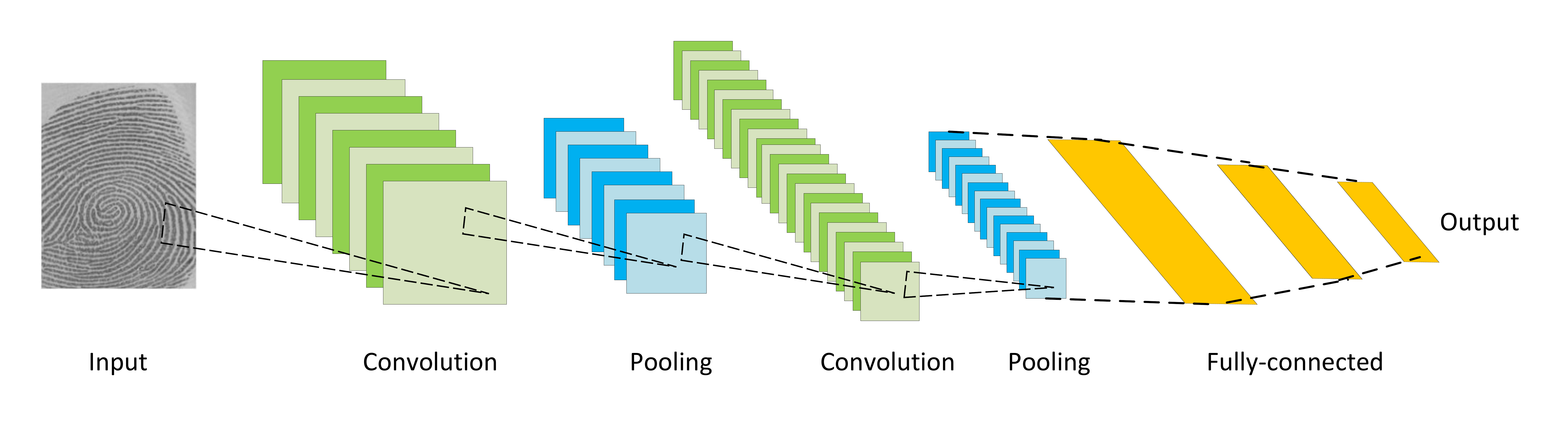}
\caption{Architecture of CNNs. From \cite{CNN}.}
\label{fig:CNN_arch}
\end{figure}

CNNs are among the most successful and widely used architectures in
the deep learning community, especially for computer vision tasks.
CNNs were initially proposed by Fukushima~\cite{neocog} in his seminal
paper on the ``Neocognitron'', which was based on Hubel and Wiesel's hierarchical receptive
field model of the visual cortex.
Subsequently, Waibel \textit{et al.}~\cite{Waibel} introduced CNNs with weights shared among temporal receptive fields and backpropagation
training for phoneme recognition, and LeCun \textit{et al.}~\cite{CNN}
developed a practical CNN architecture for document recognition
(Fig.~\ref{fig:CNN_arch}).
CNNs usually include three types of layers: i) convolutional layers,
where a kernel (or filter) of weights is convolved to extract
features; ii) nonlinear layers, which apply (usually element-wise) an activation function to
feature maps, thus enabling the network to model
nonlinear functions; and iii) pooling layers, which reduce spatial resolution by
replacing small neighborhoods in a feature map with some statistical
information about those neighborhoods (mean, max, etc.). The neuronal units in layers are locally connected; that
is, each unit receives weighted inputs from a small neighborhood,
known as the receptive field, of units in the previous layer. By
stacking layers to form multi-resolution pyramids, the higher-level
layers learn features from increasingly wider receptive fields. The
main computational advantage of CNNs is that all the receptive fields
in a layer share weights, resulting in a significantly smaller number
of parameters than fully-connected neural networks.
Some of the most well known CNN architectures include AlexNet~\cite{alexnet}, VGGNet~\cite{vggnet}, and ResNet~\cite{resnet}.

\subsection{R-CNN Based Models}
\begin{figure}[h]
\centering
\includegraphics[width=0.5\linewidth]{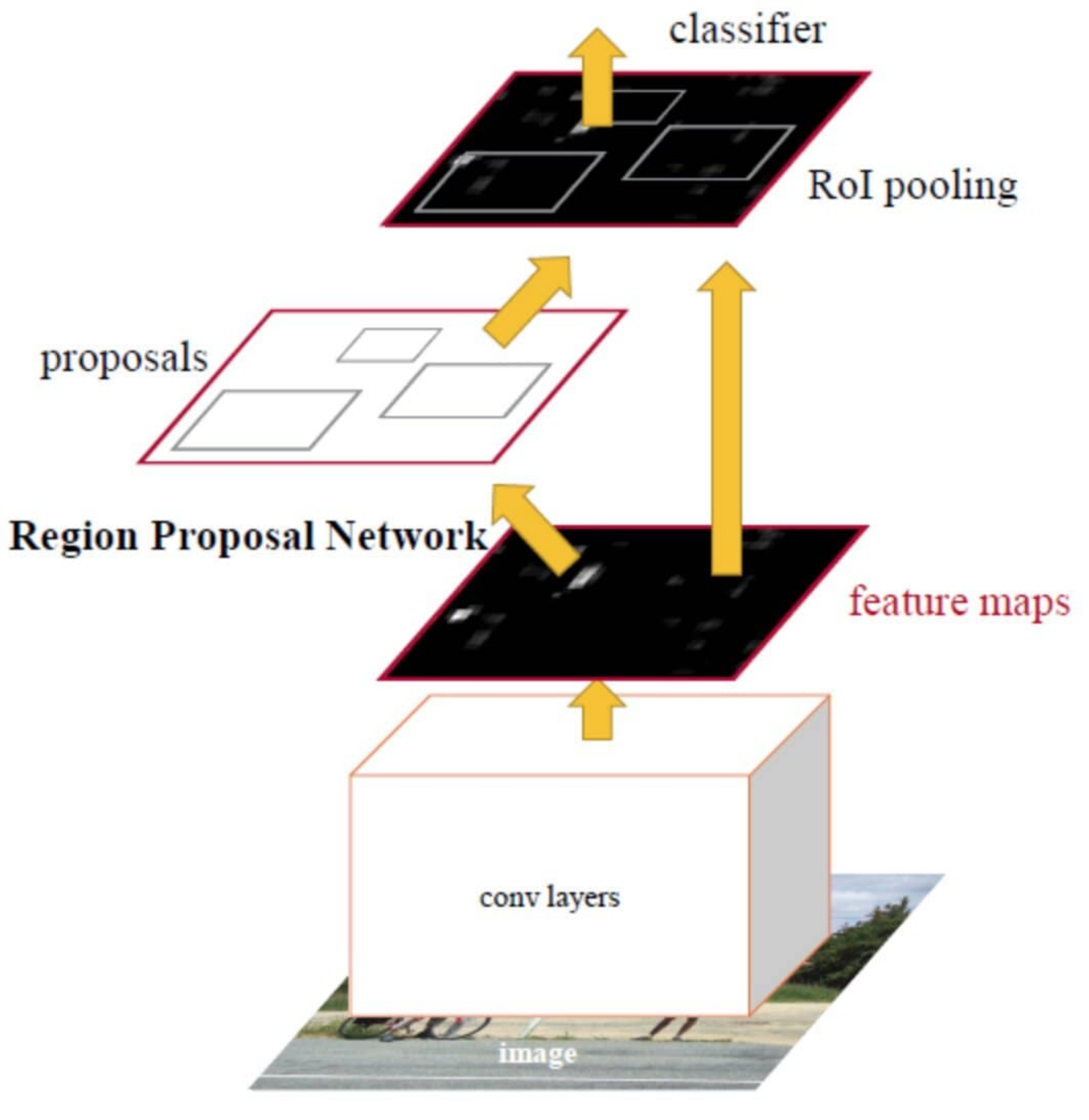}
\caption{Faster R-CNN architecture. 
Each image is processed by convolutional layers and its features are extracted, a sliding window is used in RPN for each location over the feature map, for each location, $k$ ($k=9$) anchor boxes are used (3 scales of 128, 256 and 512, and 3 aspect ratios of 1:1, 1:2, 2:1) to generate a region proposal; A cls layer outputs 2k scores to indicate whether or not there is an object for $k$ boxes; A reg layer outputs $4k$ for the coordinates (box center coordinates, width and height) of $k$ boxes.
From~\cite{ren2015faster}.}
\label{fig:faster_rcnn}
\end{figure}

The Regional CNN (R-CNN) and its extensions have proven successful in object detection
applications. 
In particular, the Faster R-CNN~\cite{ren2015faster} architecture
(Fig.~\ref{fig:faster_rcnn}) uses a
region proposal network (RPN) that proposes bounding box candidates. The
RPN extracts a Region of Interest (RoI), and an RoIPool layer computes
features from these proposals to infer the bounding box coordinates and class of the object.
Some extensions of R-CNN have been used to address the
instance segmentation problem; i.e., the task of simultaneously
performing object detection and segmentation.

\begin{figure*}[h]
\centering
\includegraphics[page=3,width=0.9\linewidth]{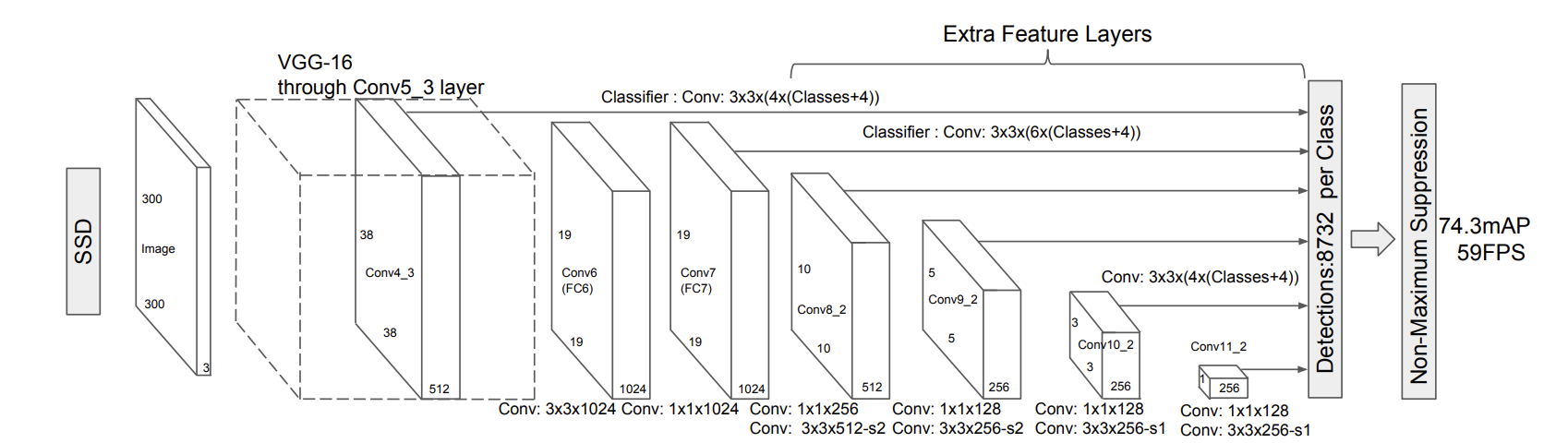}
\caption{The Architecture of SSD Model. From \cite{liu2016ssd}.}
\label{fig:SSD}
\end{figure*}

\subsection{Single Shot MultiBox Detector}
Single  Shot  Detector  (SSD)  is  a  popular  deep learning architecture proposed for object detection \cite{liu2016ssd}. It discretizes the  output  space  of  bounding  boxes  into  a  set  of  default boxes over different aspect ratios and scales per feature map location. At prediction time, the network generates scores for the presence of each object category in each default box and  produces  adjustments  to  the  box  to  better  match  the object shape. Additionally, the network combines predictions from  multiple  feature  maps  with  different  resolutions  to naturally  handle  objects  of  various  sizes.  SSD  is  simple relative  to  methods  that  require  object  proposals  because it completely eliminates proposal generation and subsequent pixel  or  feature  resampling  stages  and  encapsulates  all computation  in  a  single  network.  Figure  \ref{fig:SSD}  illustrates  the high-level architecture of the original SSD model.

\subsection{Feature Pyramid Network (FPN)}
\begin{figure}[h]
\centering
\includegraphics[page=1,width=0.99\linewidth]{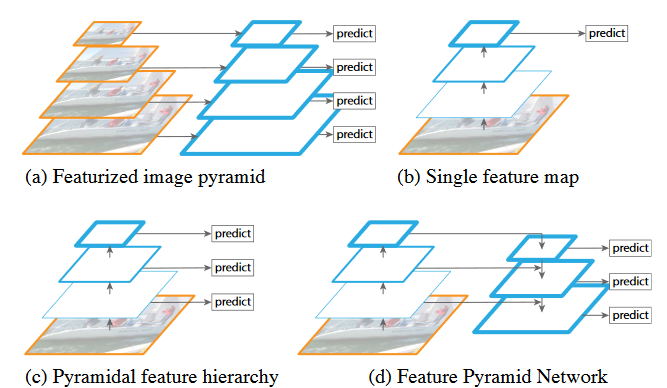}
\caption{Architecture of a simple feature pyramid network. Courtesy of \cite{lin2017feature}.}
\label{fig:fpn}
\end{figure}

Feature pyramids are a basic component in recognition systems for detecting objects at different scales \cite{lin2017feature}.
In this work, Lin et al. exploited the inherent multi-scale, pyramidal hierarchy of deep convolutional networks to construct feature pyramids with marginal extra cost.
They developed a top-down architecture with lateral connections for building high-level semantic feature maps at all scales (called FPN), and demonstrated that FPN can bring significant improvements in several vision tasks.
Fig \ref{fig:fpn} shows the high-level architecture of feature pyramid network proposed in \cite{lin2017feature}.

\begin{figure*}[t]
\centering
\includegraphics[width=0.9\linewidth]{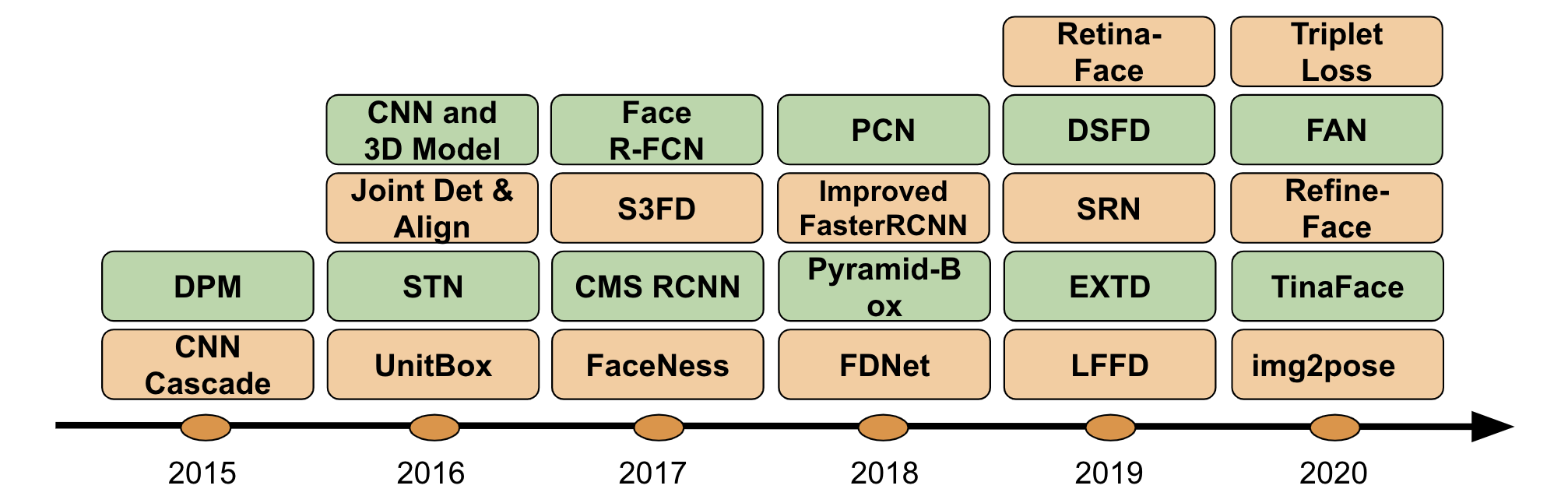}
\caption{The timeline of deep-learning-based face detection algorithms from 2015 to 2020.}
\label{fig:timeline}
\end{figure*}

\subsection{Generative Adversarial Networks (GANs)}
\begin{figure}[t]
\centering
\includegraphics[width=0.9\linewidth]{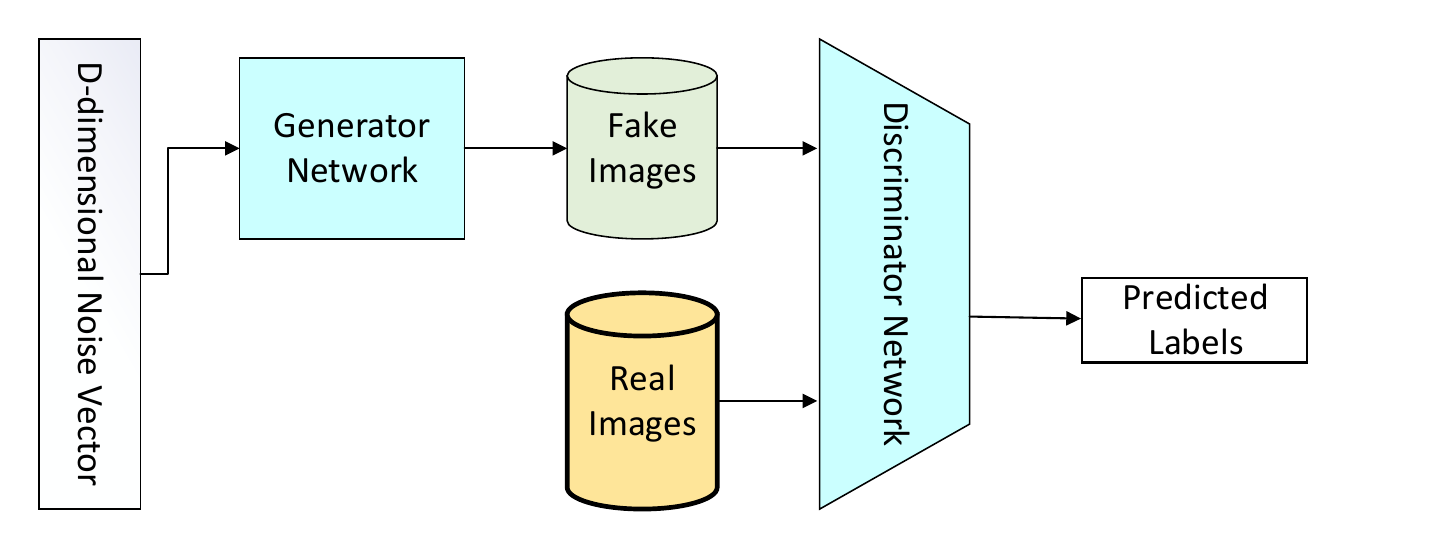}
\caption{Architecture of a GAN. Courtesy of Ian Goodfellow.}
\label{fig:gen_arch}
\end{figure}

GANs~\cite{GAN} are a newer family of deep learning models. They
consist of two networks---a generator and a discriminator
(Fig.~\ref{fig:gen_arch}). In the conventional GAN, the generator network $G$ learns a mapping from noise $z$ (with a
prior distribution) to a target distribution $y$, which is similar to
the ``real'' samples. The discriminator network $D$ attempts to
distinguish the generated ``fake'' samples from the real ones.
GAN training may be characterized as a minimax game between $G$ and $D$, where $D$ tries to minimize its classification error in distinguishing fake samples from real ones,
hence maximizing a loss function, and $G$ tries to maximize the
discriminator network's error, hence minimizing the loss function.
Some of the GAN variants include Convolutional-GANs~\cite{radford2015unsupervised}, conditional-GANs~\cite{mirza2014conditional}, Wasserstein-GANs~\cite{arjovsky2017wasserstein}, and CycleGAN \cite{zhu2017unpaired}.

\section{Overview of Different Deep Face Detection Models}
\label{sec:DL_models}
There are many approaches proposed for face detection using different deep learning architectures. 
In order to better summarize the existing works in a more comprehensive way, we grouped these works into a few prominent categories and review the major works of each category in the below sections:
\begin{itemize}
    \item Cascade-CNN Based Models
    \item R-CNN Based Models
    \item Single Shot Detector Models
    \item Feature Pyramid Network Based Models
    \item Transformers Based Models
    \item Other Architectures
\end{itemize}

Figure \ref{fig:timeline} provides an illustration summarizing the most popular deep learning based face detection models from 2015 till 2021.


\subsection{Cascade-CNN Based Models}
\begin{figure}[h]
\centering
\includegraphics[width=0.99\linewidth]{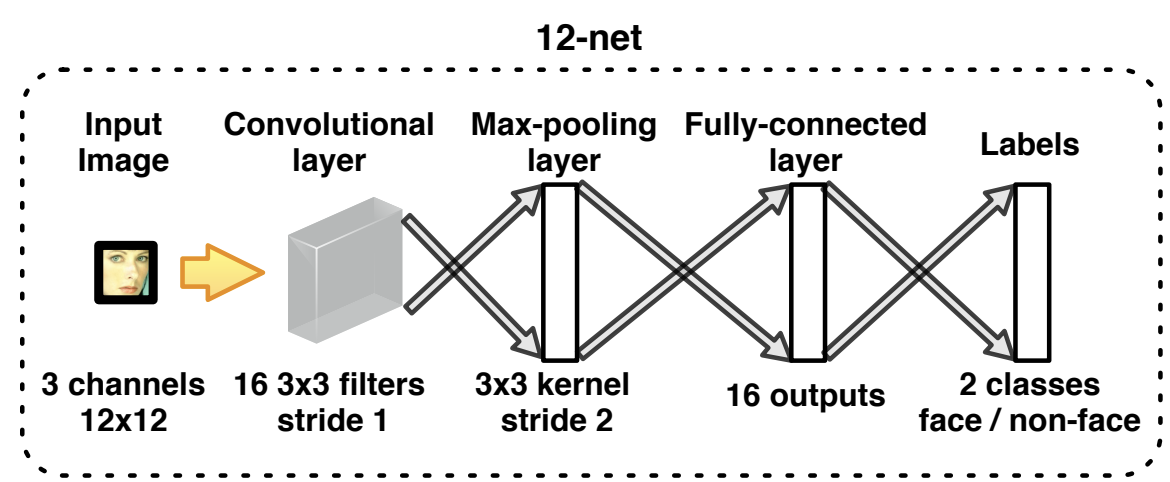}
\caption{The architecture of 12-layer Cascade-CNN model. Courtesy of \cite{li2015convolutional}.}
\label{fig:cascade_cnn}
\end{figure}

Li et al. \cite{li2015convolutional} proposed one of the early deep models for face detection, based on a convolutional neural network cascade.
The proposed CNN cascade operates at multiple resolutions, quickly rejects the background regions in the fast low resolution stages, and carefully evaluates a small number of 
candidates in the last high resolution stage.
To improve localization effectiveness, and reduce the number of candidates at later stages, they introduce a CNN-based calibration stage after each of the detection stages in the cascade.
The proposed method runs at 14 FPS on a
single CPU core for VGA-resolution images and 100 FPS using a GPU.
The architecture of a 12-layer CNN-Cascade network is shown Figure \ref{fig:cascade_cnn}.

\begin{figure}[h]
\centering
\includegraphics[width=0.8\linewidth]{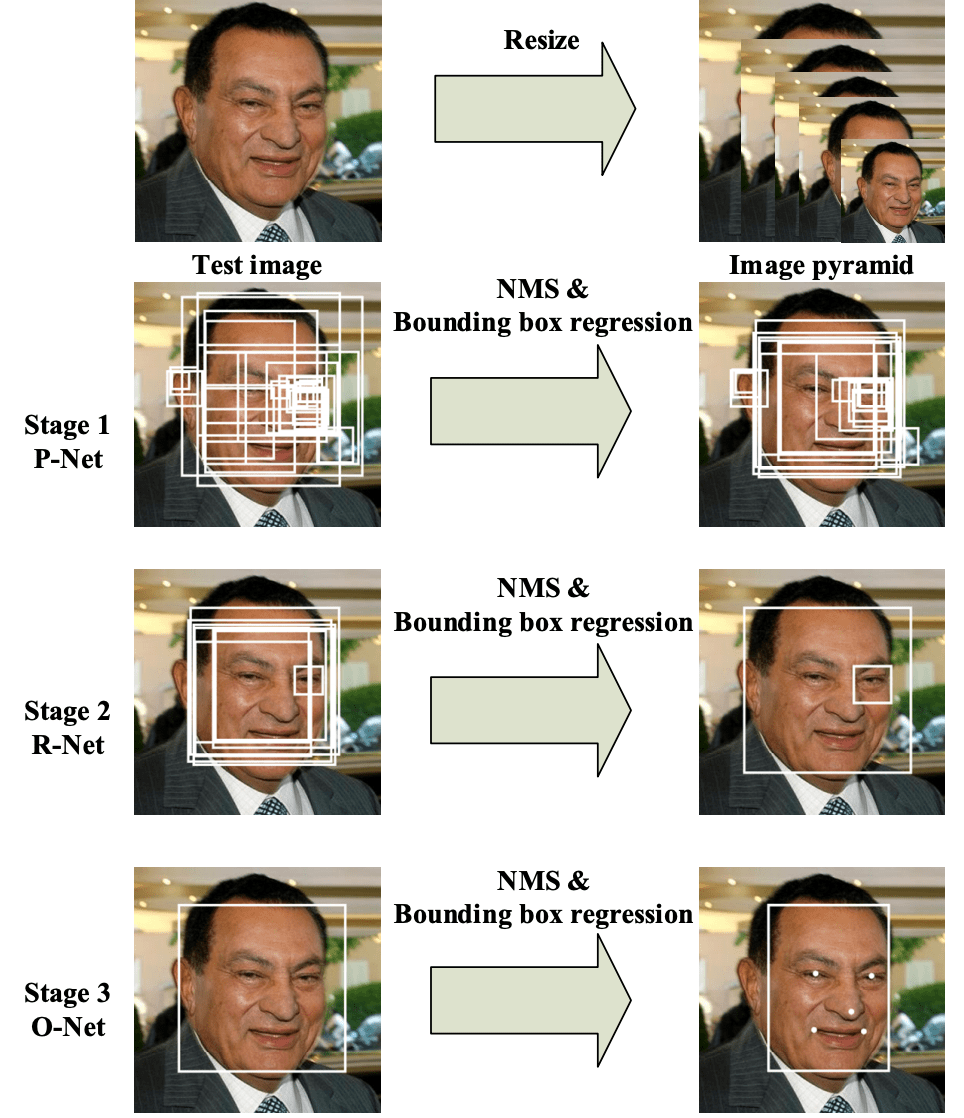}
\caption{The architecture of the proposed multi-task cascaded CNN. Courtesy of \cite{zhang2016joint}}
\label{fig:mt_cascade}
\end{figure}

This approach was 
extended by Zhang et al. \cite{zhang2016joint} for joint face detection and alignment.
This work leverages a cascaded architecture with three stages of carefully designed deep convolutional networks to predict face and landmark location in a coarse-to-fine manner.
They also proposed a new online hard sample mining strategy to further improve
accuracy.
The high-level architecture of this model is shown in Fig \ref{fig:mt_cascade}.

\begin{figure*}[h]
\centering
\includegraphics[width=0.8\linewidth]{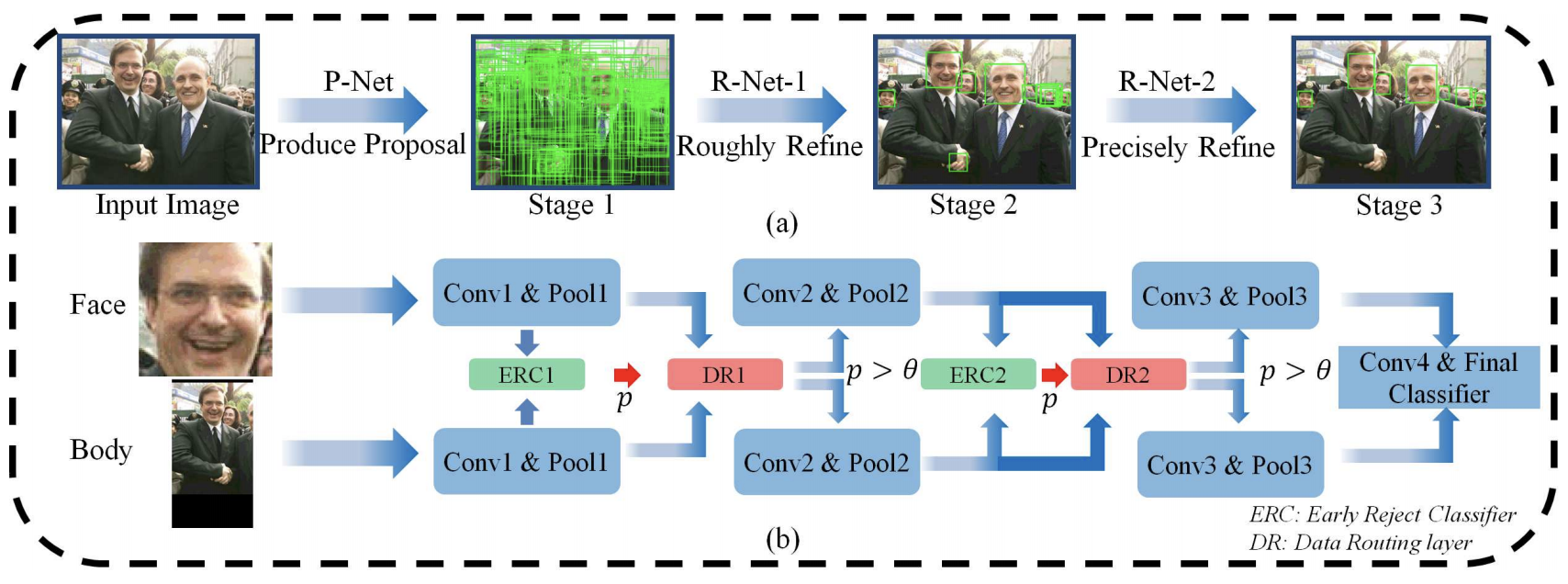}
\caption{The architecture of the proposed Inside Cascaded CNN. Courtesy of \cite{zhang2017detecting}}
\label{fig:ins_cascade}
\end{figure*}

In a subsequent work, Zhang et al. \cite{zhang2017detecting} proposed "Inside Cascaded Structure" that introduces face/non-face classifiers at different layers within the same CNN.
They introduce a two-stream contextual CNN architecture that leverages body part information adaptively to enhance face
detection.
In the training phase, they proposed a data routing mechanism which enables different layers to be trained by different types of samples, and thus deeper layers can focus on handling more difficult samples compared with traditional architecture.
The architecture of this model is shown in Fig \ref{fig:ins_cascade}.

There are other follow-up works that also rely on cascaded CNN architectures, including "Joint training of cascaded CNN for face detection" \cite{qin2016joint}, and "Face detection method based on cascaded convolutional networks" \cite{qi2019face}.
%
%

\subsection{R-CNN and Faster-RCNN Based Models}

Region proposal-based CNN models have been very successful for object detection, and 
have also been applied to face detection by several works.
In \cite{chen2016supervised}, Chen et al. proposed a cascaded CNN model for face detection, called Supervised Transformer Network (STN).
The first stage is a multi-task Region Proposal Network (RPN), which simultaneously predicts candidate face regions
along with associated facial landmarks. The candidate regions are then warped by mapping the detected facial landmarks to their canonical positions to better normalize the face patterns. The second stage, which is a R-CNN, then verifies if the warped candidate regions are valid faces or not.

Jiang and Miller \cite{jiang2017face} explored the application of faster-RCNN model \cite{ren2015faster} on face detection. They were able to achieve state-of-the-art results on two widely used face detection benchmarks, by simply training faster-RCNN on large-scale Wider-Face dataset.
This illustrates the continuing importance of the volume of training data.

\begin{figure}[h]
\centering
\includegraphics[width=0.99\linewidth]{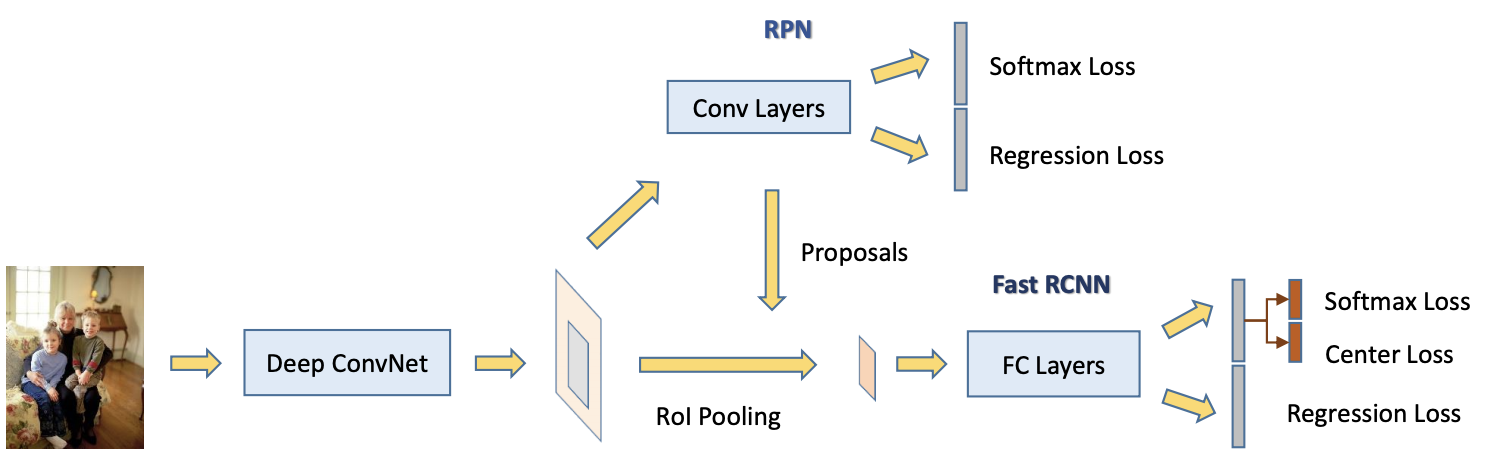}
\caption{The architecture of the Face-RCNN Model. Courtesy of \cite{wang2017face}}
\label{fig:face_rcnn}
\end{figure}

In parallel, \cite{wang2017face}, Wang et al. proposed "Face R-CNN", which applied Faster-RCNN with a few other techniques to face detection. 
They exploited several new techniques including a multi-task loss function design, online hard example mining scheme, and a multi-scale training strategy to improve Faster R-CNN in multiple aspects.
The architecture of this work is similar to Faster-RCNN, and is shown in Fig. \ref{fig:face_rcnn}.

In \cite{sun2018face}, Sun et al. proposed a face detection framework using an improved Faster-RCNN.
They improved the Faster RCNN framework by combining a number of strategies, including feature concatenation, hard negative mining, multi-scale training, model pre-training, and 
careful calibration of key parameters.

\begin{figure}[h]
\centering
\includegraphics[width=0.8\linewidth]{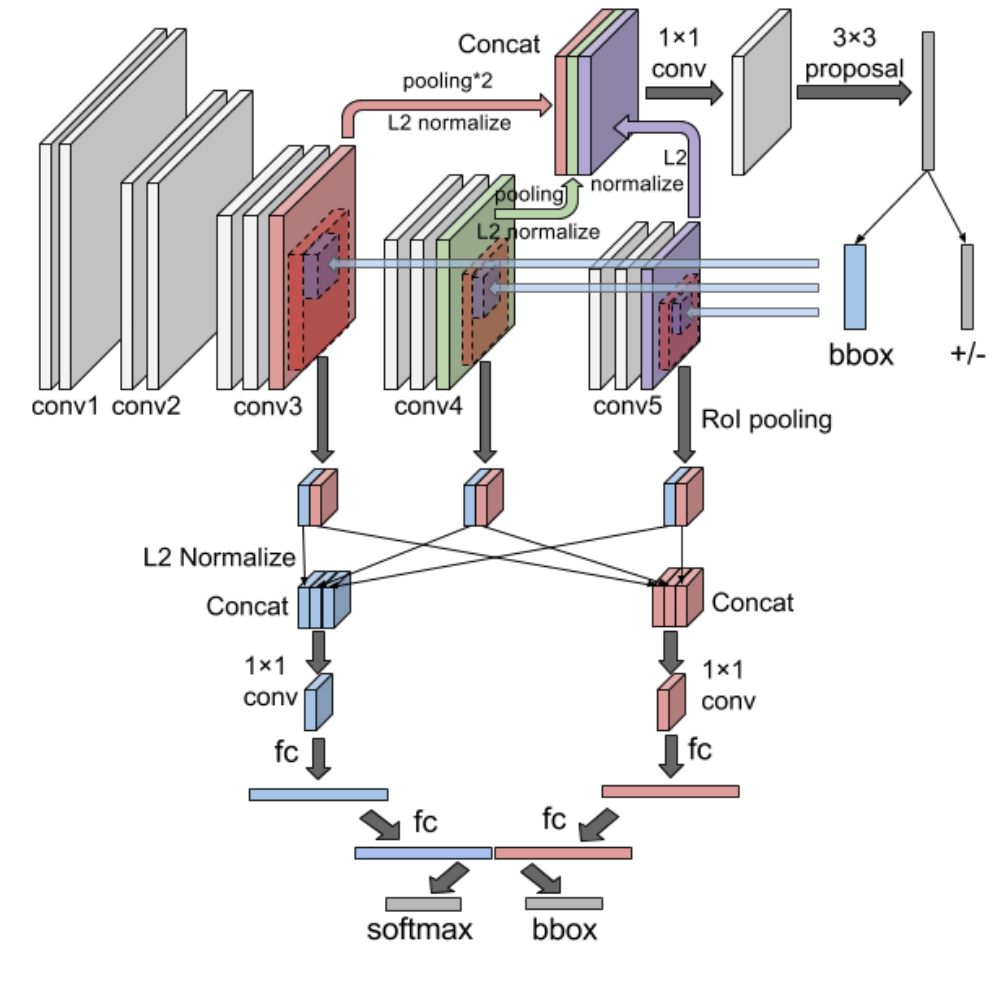}
\caption{The architecture of the CMS-RCNN Model. Courtesy of \cite{zhu2017cms}}
\label{fig:CMS_CNN}
\end{figure}

To better incorporate contextual information, Zhu et al. \cite{zhu2017cms} proposed "CMS-RCNN", which is a contextual multi-scale region-based CNN model for unconstrained
face detection.
In addition to the typical region proposal component and the region-of-interest (RoI) detection, they introduce two main ideas specifically for face detection. First, the multi-scale information is grouped both in region proposal and RoI detection to deal with tiny face regions. Second, the proposed network incorporates explicit body contextual reasoning in the network, inspired by the human vision system.
Fig. \ref{fig:CMS_CNN} illustrates the architecture of the CMS-RCNN.

\begin{figure}[h]
\centering
\includegraphics[width=0.99\linewidth]{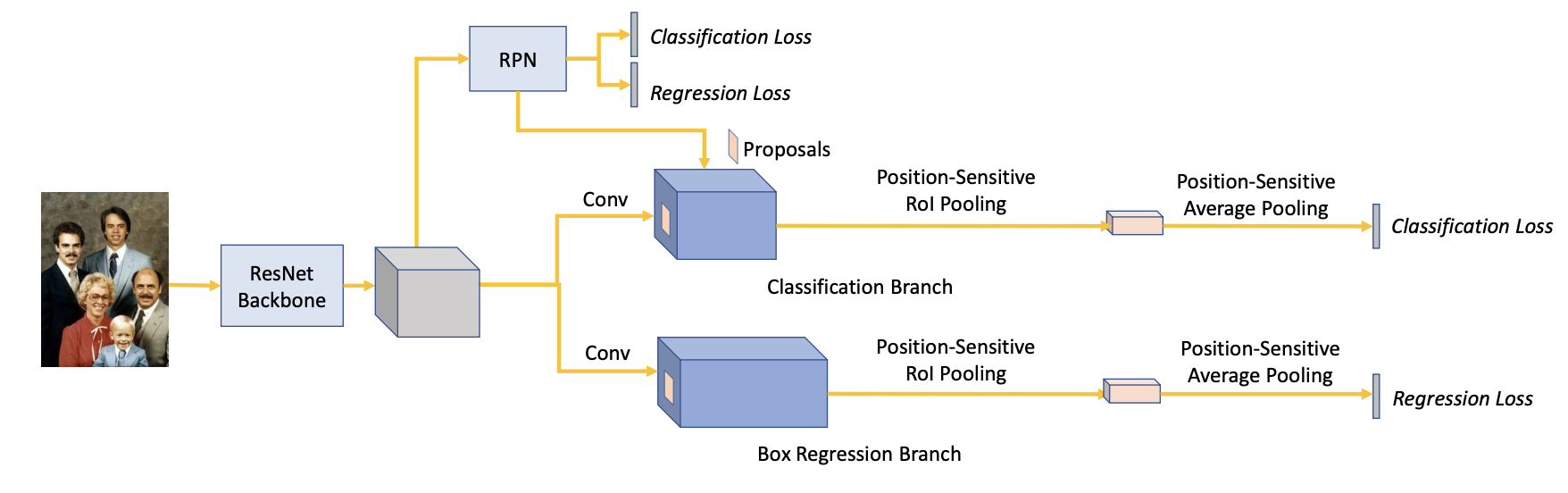}
\caption{The architecture of the R-FCN Model. Courtesy of \cite{wang2017detecting}}
\label{fig:R-FCN}
\end{figure}

In another effort, \cite{wang2017detecting}, Wang et al. proposed a region-based fully
convolutional network for face detection.
They adopt a fully convolutional Residual Network (ResNet) as the backbone network, and exploit several new techniques including
position-sensitive average pooling, 
%
%
multi-scale training and testing and on-line
hard example mining strategy to improve the detection accuracy.
Fig. \ref{fig:R-FCN} illustrates the architecture of the proposed R-FCN model.

Some other R-CNN based face detection methods include: "Face detection with different scales based on faster R-CNN" \cite{wu2018face}, "  Face Detection Using Improved Faster RCNN" \cite{zhang2018face} , and "Design of a Deep Face Detector by Mask R-CNN" \cite{cakiroglu2019design}.
%
%

\subsection{Single Shot Detection Models}
Single stage detection (SSD) is another popular and major direction in deep learning based face detection. 
Unlike two-stage proposal-classification detectors, such as R-CNN models, SSD detects faces in a single stage directly from the early
convolutional layers in a classification network.

\begin{figure*}[h]
\centering
\includegraphics[width=0.75\linewidth]{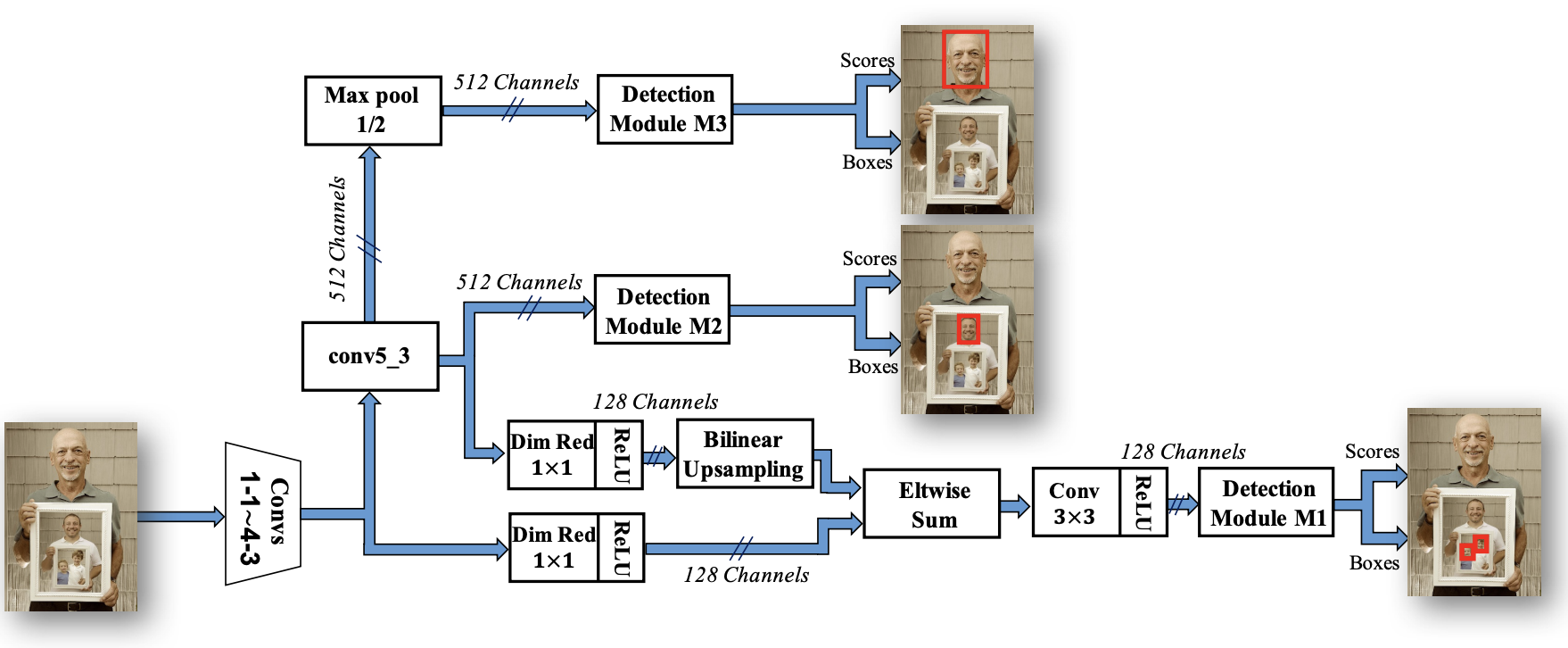}
\caption{The architecture of the proposed SSH Model. Courtesy of \cite{najibi2017ssh}}
\label{fig:SSH}
\end{figure*}

In \cite{najibi2017ssh}, Najibi et al. proposed a " Single Stage Headless (SSH)" face detection framework, which achieved state of the art results on all subsets of Wider Faces, FDDB, and Pascal-Faces. 
Instead of relying on an image pyramid to detect faces with various scales, SSH is scale-invariant by design. It simultaneously detects faces with different scales in a single forward pass of the network, but from different layers.
These properties make SSH fast and light-weight.
Fig. \ref{fig:SSH} illustrates the architecture of the proposed SSH model.

\begin{figure}[h]
\centering
\includegraphics[width=0.99\linewidth]{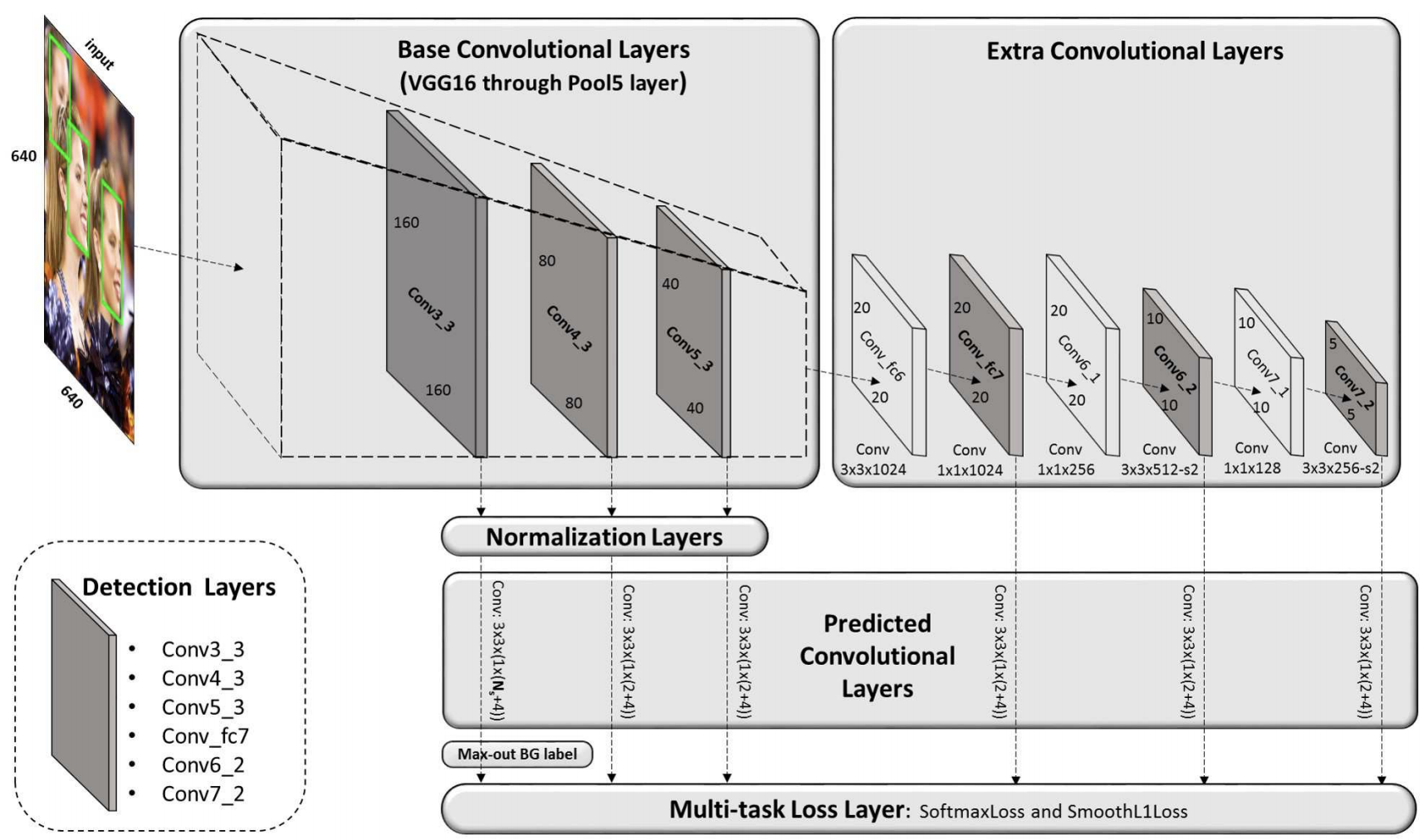}
\caption{The architecture of the proposed S3FD Model. Courtesy of \cite{najibi2017ssh}}
\label{fig:s3fd}
\end{figure}

Zhang et al. proposed "Single Shot Scale-invariant Face Detector (S$^3$FD)" \cite{zhang2017s3fd}, to address scale variations better with a single
deep neural network.
Detection of  for small faces is an especially common problem with anchor-based detectors.
There are three main contributions in this work. 
First,  a scale-equitable face detection framework is proposed to handle different scales of faces well.
Second, the recall rate of small faces is improved by a scale compensation anchor-matching strategy.
Third, the false positive rate of small faces is reduced via a max-out background label.
Fig. \ref{fig:s3fd} illustrates the architecture of the proposed S3FD model.

Zhang et al. proposed FaceBoxes \cite{zhang2017faceboxes}, which is a real-time face detector based on the "Rapidly Digested Convolutional Layers (RDCL)" and the "Multiple Scale Convolutional
Layers (MSCL)".
RDCL is designed to enable FaceBoxes to achieve real-time speed.
The MSCL
aims at enriching the receptive fields and discretizing anchors over different layers to handle faces of various scales.
They also propose a new anchor densification strategy to make different types of anchors have the same density on the image.
This significantly improves the recall rate
of small faces.
One interesting fact about FaceBoxes is that its speed is invariant to the number of faces.

\begin{figure}[h]
\centering
\includegraphics[width=0.99\linewidth]{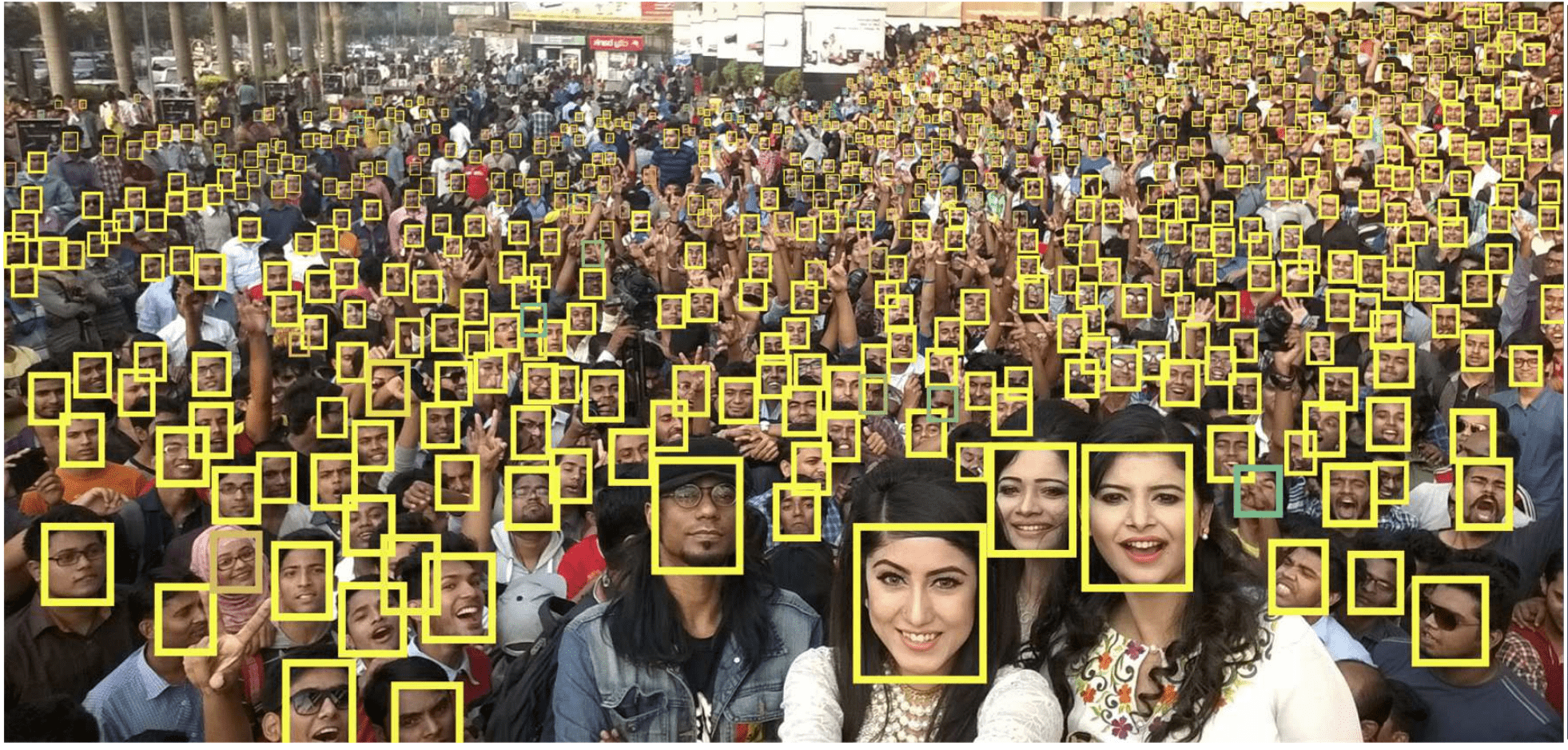}
\caption{The detected faces by the tiny face detector model for a sample image. Courtesy of \cite{hu2017finding}}
\label{fig:hu_sample}
\end{figure}

\begin{figure}[h]
\centering
\includegraphics[width=0.99\linewidth]{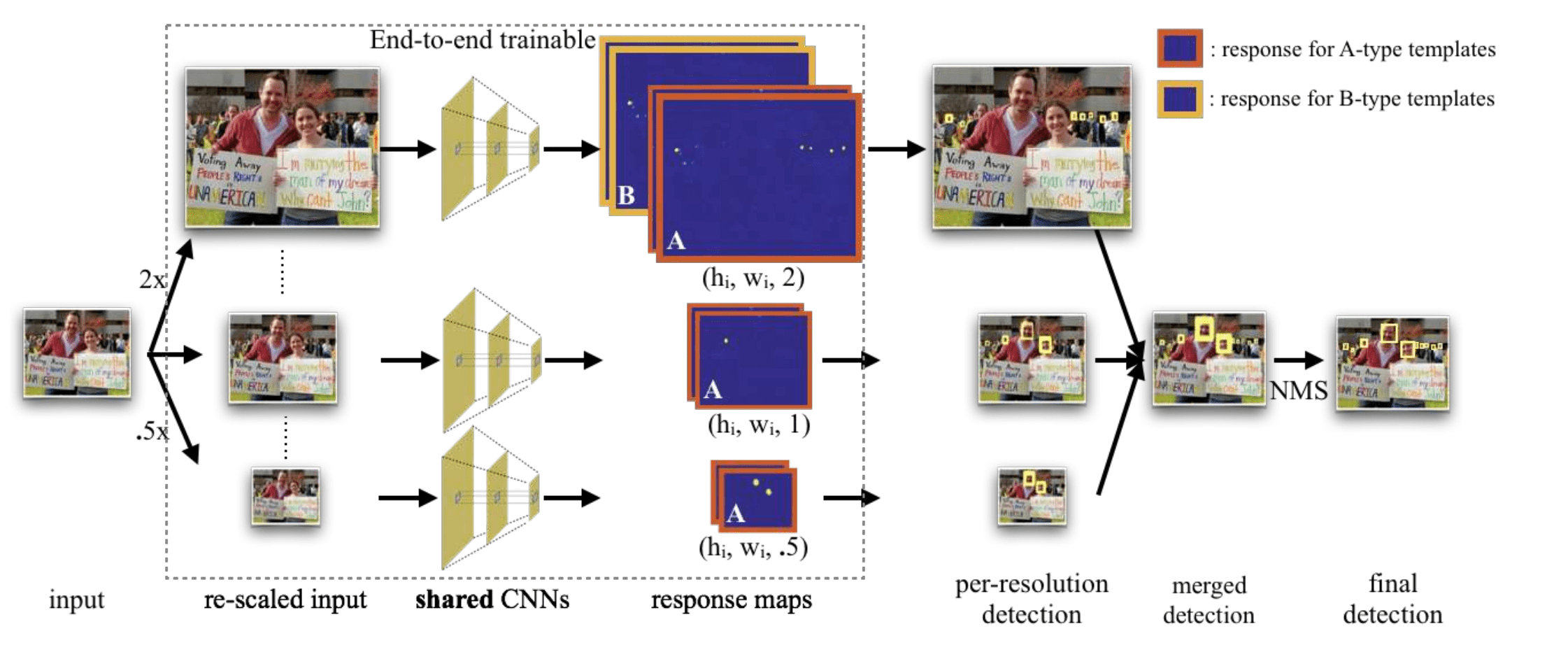}
\caption{The overall architecture of the model for detecting tiny faces. Courtesy of \cite{hu2017finding}}
\label{fig:hu_arch}
\end{figure}

Hu and  Ramanam proposed an interesting algorithm designed to address the problem with detecting small faces \cite{hu2017finding}. 
This model is able to detect a few hundred of small faces in a single image. 
One example of such a case is shown in Fig \ref{fig:hu_sample}.
They explored three aspects of the problem of finding small faces:  the role of scale invariance, image resolution, and contextual reasoning.
They took a different approach from previous works, and trained separate detectors for different scales.
To maintain efficiency, detectors are trained in a multi-task fashion: they make use of features extracted from multiple layers of single (deep) feature hierarchy. 
They show that the context is very important and make use of massively-large receptive fields.
The overall architecture of this model is shown in Fig \ref{fig:hu_arch}.

\begin{figure}[h]
\centering
\includegraphics[width=0.99\linewidth]{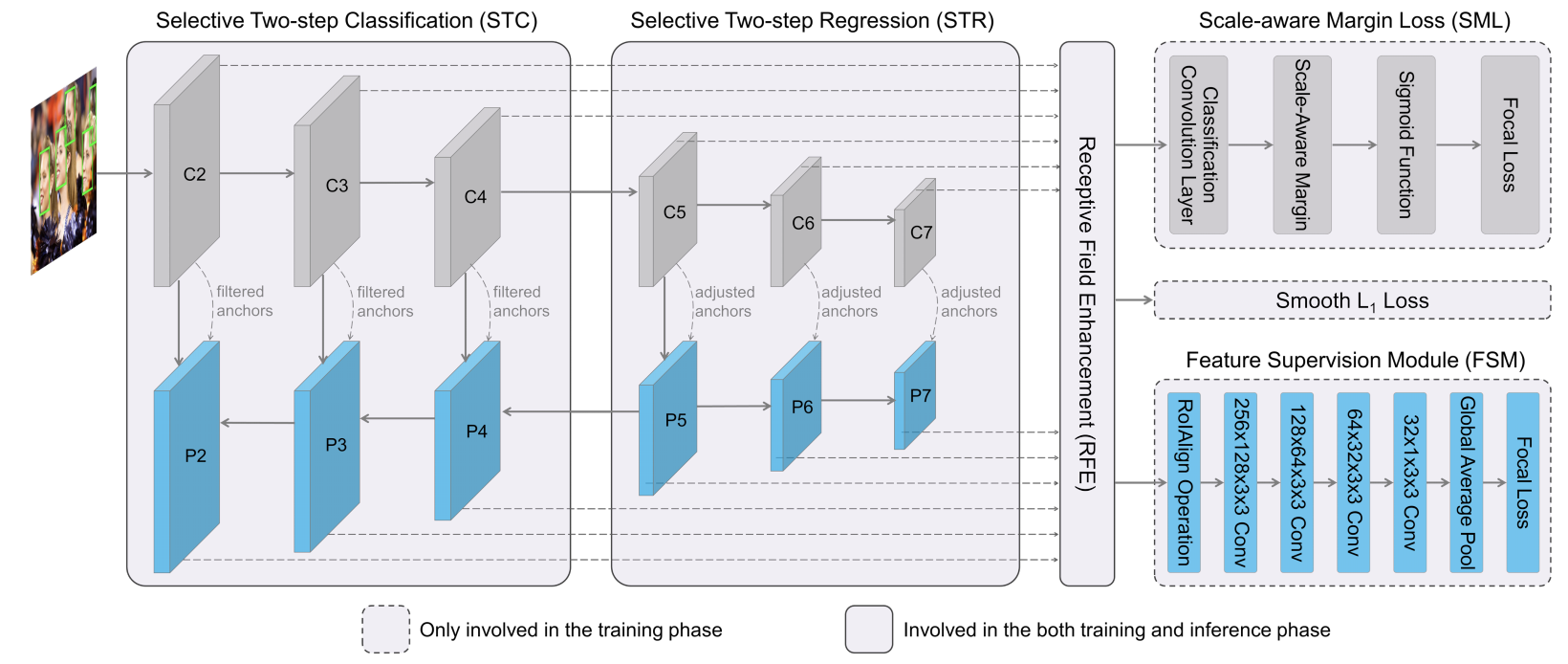}
\caption{The overall architecture of the RefineFace model for detecting faces. Courtesy of \cite{zhang2020refineface}}
\label{fig:refine_face}
\end{figure}

Most recently, in \cite{zhang2020refineface}, Zhang et al. proposed  a single-shot refinement face detector called RefineFace.
This framework is based on RetinaNet \cite{lin2017focal}, with five proposed modules: Selective Two-step Regression (STR), Selective Two-step Classification (STC), Scale-aware Margin Loss (SML), Feature Supervision Module (FSM) and Receptive Field Enhancement (RFE).
To enhance the regression ability for high location accuracy, STR coarsely adjusts locations and sizes of anchors from high level detection layers to provide better initialization for subsequent regressors.
The overall architecture of this model is shown in Fig \ref{fig:refine_face}.

There are other works in this category, e.g. scale-friendly faces \cite{yang2017face}, YOLO-face: a real-time face detector \cite{chen2020yolo}, and "SANet: Smoothed attention network for single stage face detector" \cite{shi2019sanet}.

\subsection{Feature Pyramid Network Based Models}
Another popular family of deep models in face detection is based on feature pyramid networks \cite{lin2017feature}.
Feature pyramid networks have also been used for object detection, as well as semantic segmentation.
A feature pyramid is a neural network structure which
combines semantically weak features with semantically strong features using skip-connections.

In \cite{zhang2020feature}, Zhang et al. proposed "Feature Agglomeration Networks (FANet)", inspired by the Feature Pyramid Network, for single-stage face detection.
The key idea of this framework is to exploit the inherent multi-scale features of a single
convolutional neural network by aggregating higher-level semantic feature maps of different scales as contextual cues to augment lower-level feature maps via a hierarchical agglomeration manner at marginal extra computation cost. They also proposed a Hierarchical Loss to effectively train
the FANet model.
The overall architecture of FANet model is shown in Fig \ref{fig:fanet}.
\begin{figure}[h]
\centering
\includegraphics[width=0.99\linewidth]{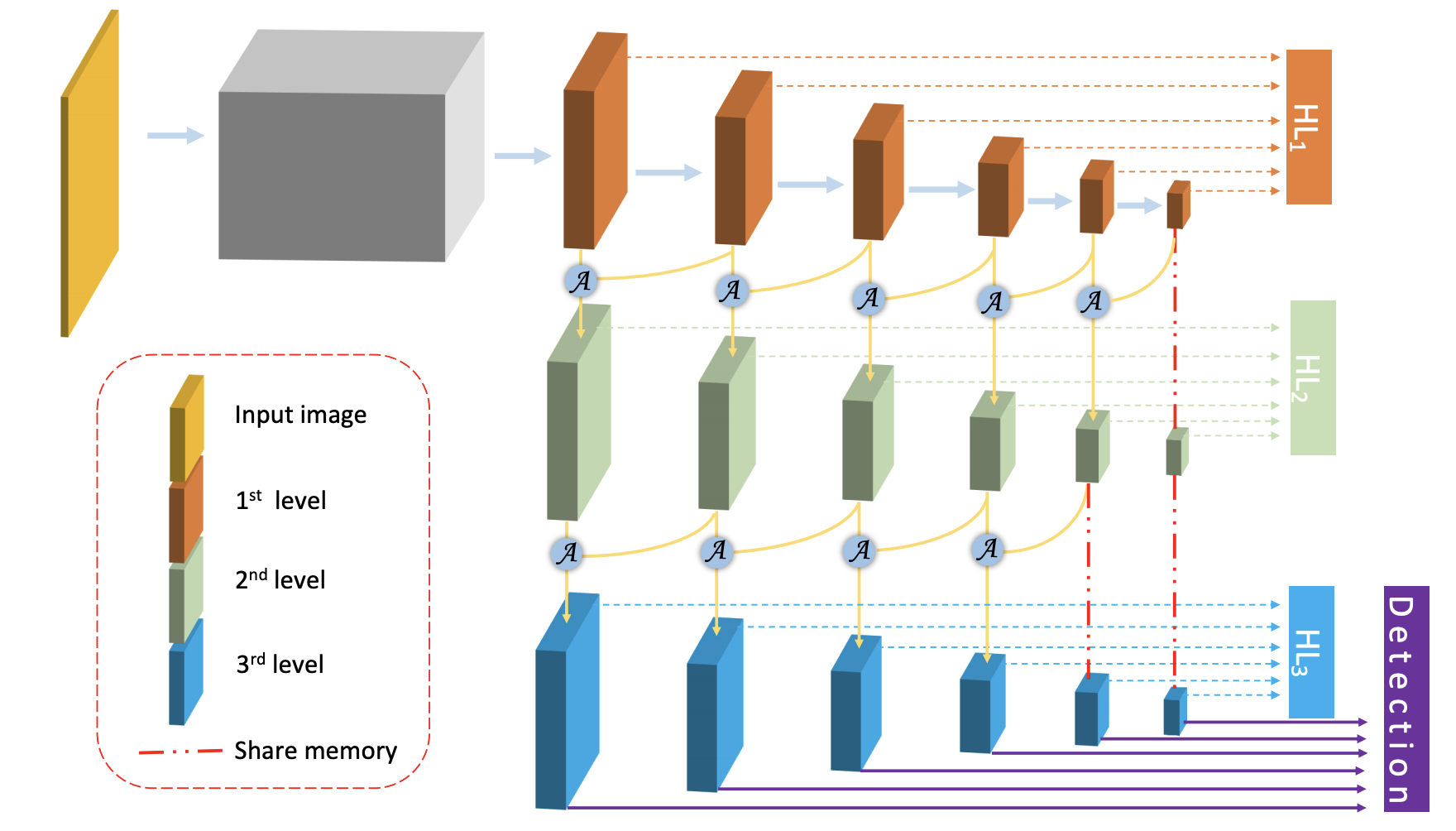}
\caption{The overall architecture of FANet model for face detection. Courtesy of \cite{zhang2020feature}}
\label{fig:fanet}
\end{figure}

\begin{figure}[h]
\centering
\includegraphics[width=0.99\linewidth]{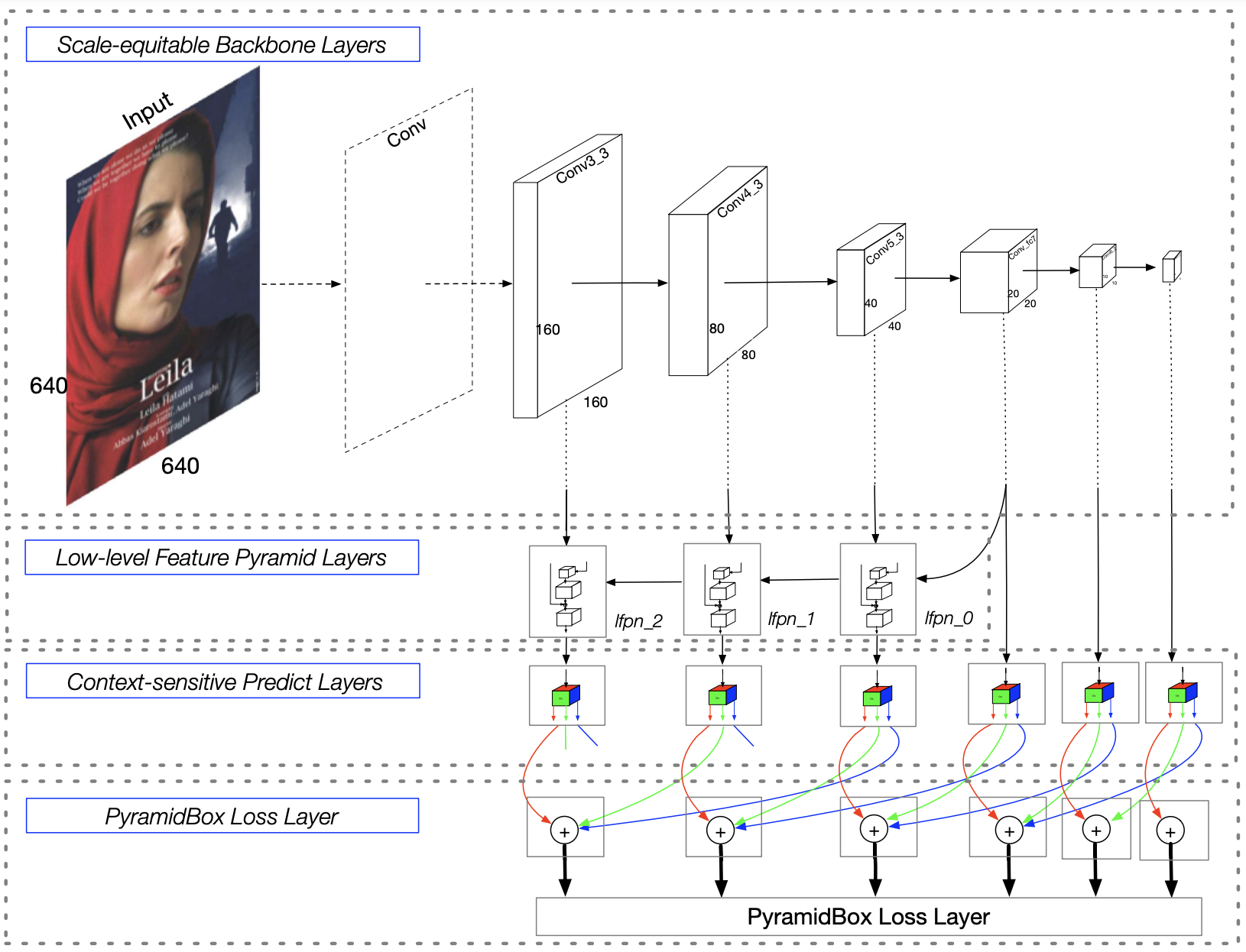}
\caption{The overall architecture of PyramidBox model for face detection. Courtesy of \cite{tang2018pyramidbox}}
\label{fig:PyramidBox}
\end{figure}

In another promising work, Tang et al. proposed PyramidBox \cite{tang2018pyramidbox}, which is a context-assisted single shot face detector, and tries to address the challenge to detect small, blurred and partially occluded faces in  uncontrolled environments.
They improved the utilization of contextual information in the following three
aspects:
First, they designed a novel context anchor to supervise high-level contextual feature learning by a semi-supervised method, which they called PyramidAnchors. 
Second, they proposed a low-level FPN to combine adequate high-level context semantic feature and low-level facial feature together, which also allows the PyramidBox to predict faces of all scales in a single shot. 
Third, they introduced a context-sensitive structure to increase the capacity of prediction network to improve the final accuracy of output.
The architecture of the proposed PyramidBox model is shown in  Fig \ref{fig:PyramidBox}.

\begin{figure}[h]
\centering
\includegraphics[width=0.99\linewidth]{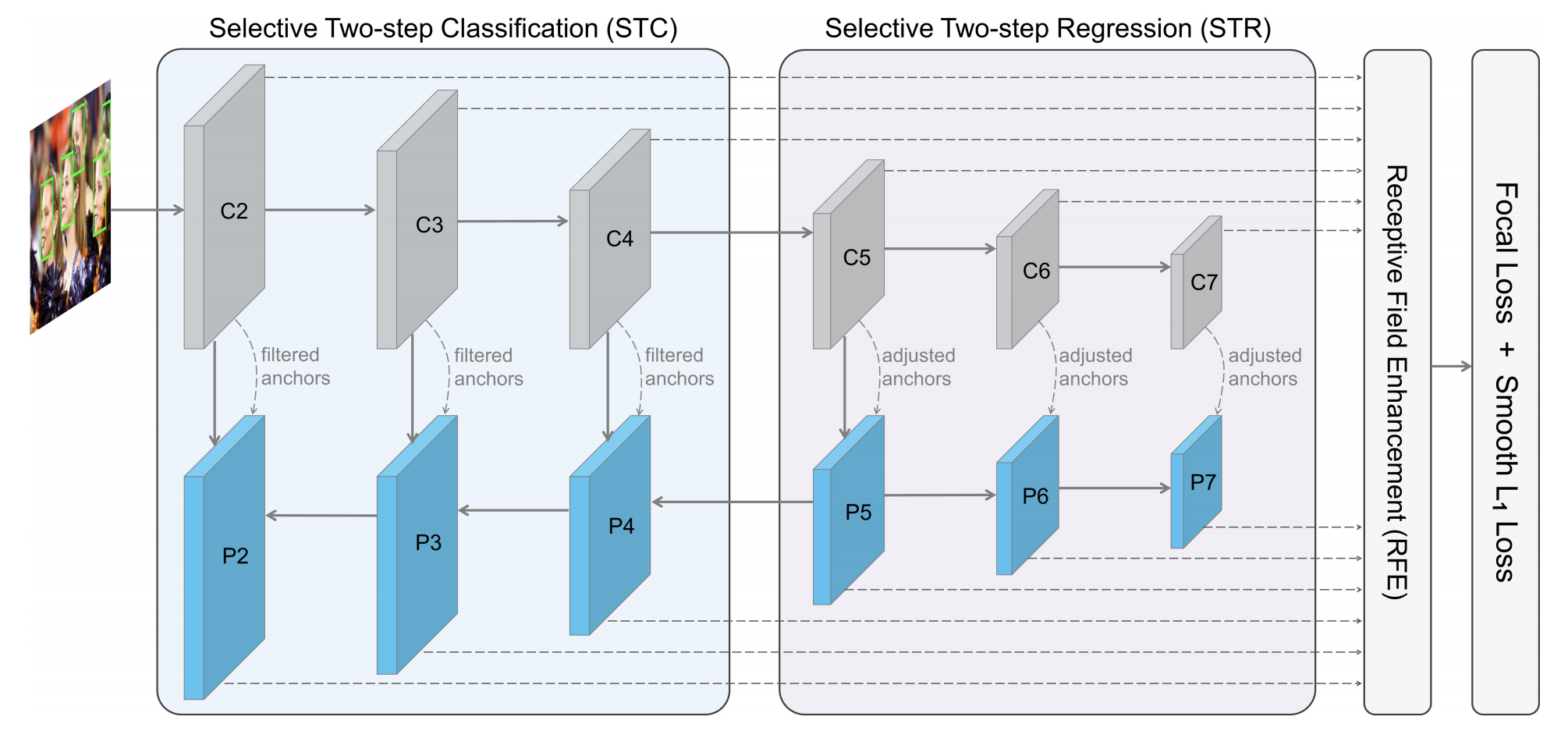}
\caption{The model architecture of Selective Refinement Network for face detection. Courtesy of \cite{chi2019selective}}
\label{fig:srn}
\end{figure}

In \cite{chi2019selective}, Chi et al. proposed a new single-shot model for face detection called "Selective Refinement Network (SRN)".
SRN consists of two novel modules: the Selective Two-step Classification (STC) module and the Selective Two-step Regression
(STR) module. The STC aims to filter out most simple negative anchors from low level detection layers to reduce the search space for the subsequent classifier, while the STR is
designed to coarsely adjust the locations and sizes of anchors from high level detection layers to provide better initialization
for the subsequent regressor.
Furthermore, they also designed a Receptive Field Enhancement (RFE) block to provide a more diverse receptive field, which helps to better capture faces in some extreme poses.
The network architecture of the proposed Selective Refinement Network model is shown in Fig \ref{fig:srn}.

\begin{figure}[h]
\centering
\includegraphics[width=0.99\linewidth]{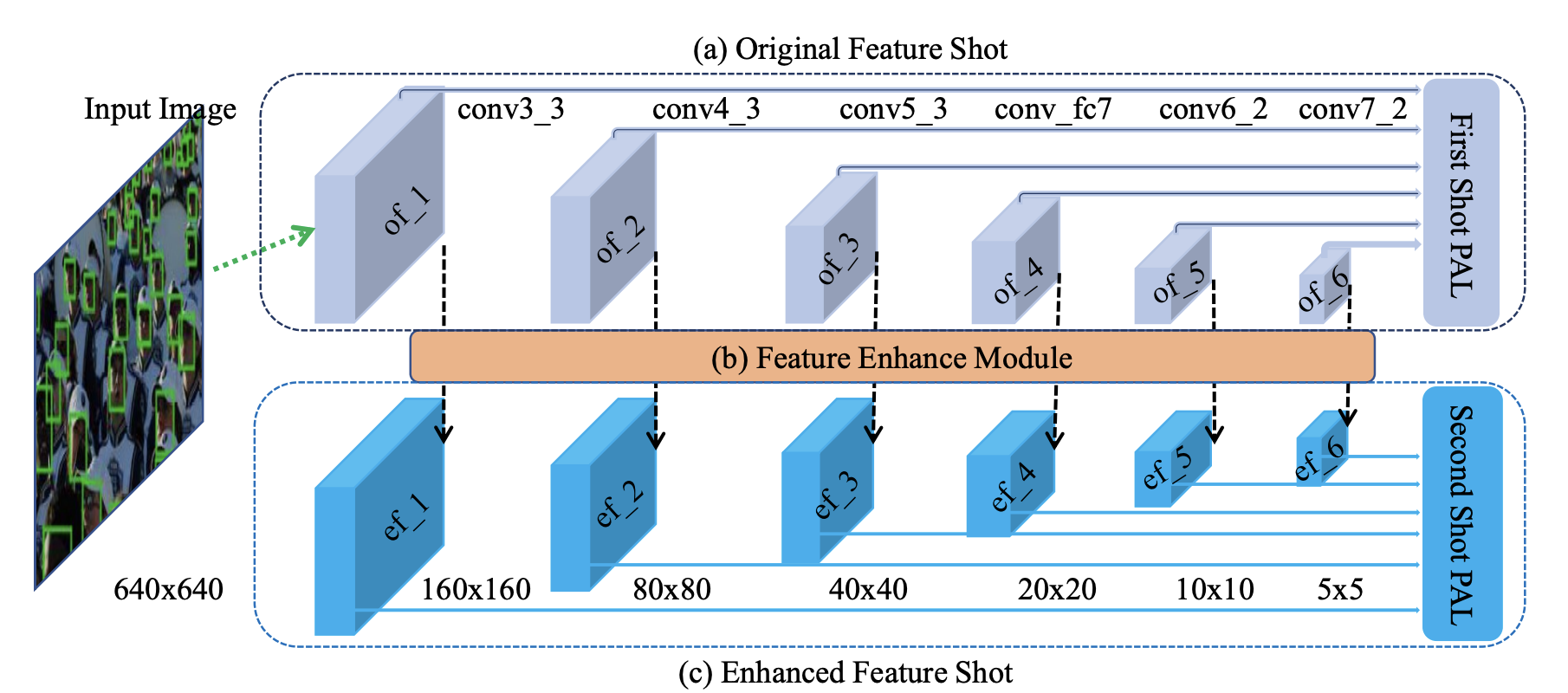}
\caption{The high-level architecture of dual shot face detector model. Courtesy of \cite{li2019dsfd}}
\label{fig:dsfd}
\end{figure}

\begin{figure}[h]
\centering
\includegraphics[width=0.99\linewidth]{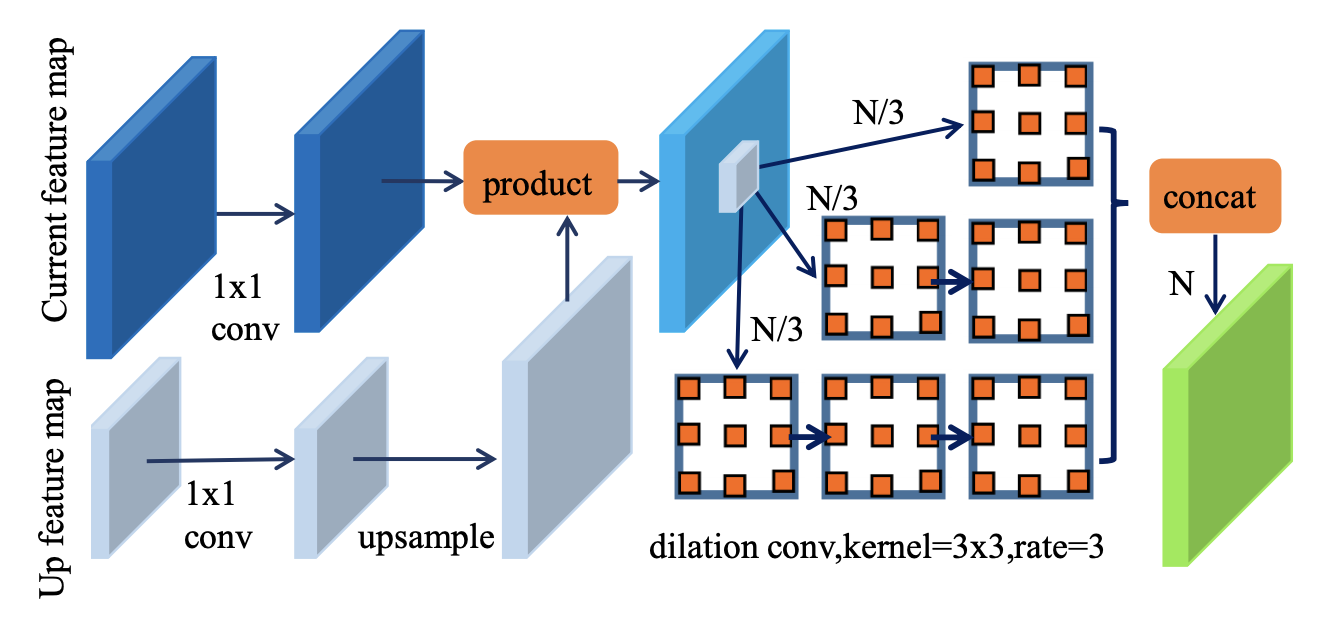}
\caption{The details of "Feature Enhance Module", used in DSFD framework. Courtesy of \cite{li2019dsfd}}
\label{fig:FEM}
\end{figure}

In \cite{li2019dsfd}, Li et al. proposed a novel face detection algorithm called "dual shot face detector (DSFD)", with three key contributions: 
First, they developed a Feature Enhance
Module (FEM) for enhancing the original feature maps to extend the single shot detector to dual shot detector. 
Second, they adopted Progressive Anchor Loss (PAL) computed by two different sets of anchors to effectively facilitate the features. 
And third, they used an Improved Anchor Matching (IAM) by integrating novel anchor assign strategy into data augmentation to provide better initialization for the regressor.
Fig \ref{fig:dsfd} provides the overall architecture of DSFD framework. 
The "Feature Enhance Module", which is a key component of this model, is shown in Fig \ref{fig:FEM}.
In \cite{li2019pyramidbox++}, Li et al. proposed PyramidBox++, which is an improved version their earlier PyramidBox framework.

\begin{figure}[h]
\centering
\includegraphics[width=0.99\linewidth]{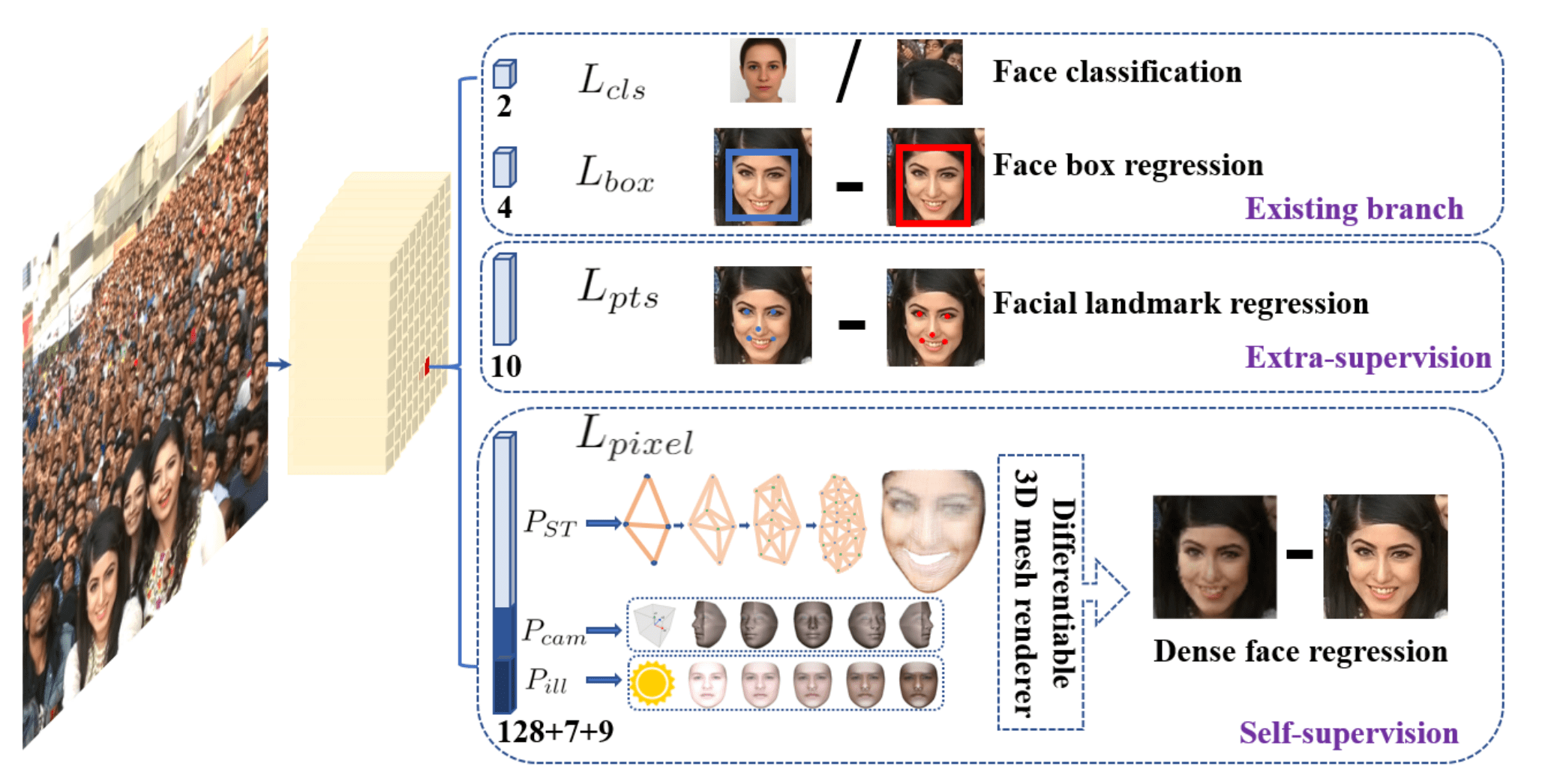}
\caption{The architecture of RetinaFace framework for face detection. Courtesy of \cite{deng2019retinaface}}
\label{fig:retina}
\end{figure}

In \cite{deng2019retinaface}, Dent et al. proposed a very popular single-stage face detection model of this category, which is called RetinaFace \cite{deng2019retinaface}.
RetinaFace performs pixel-wise face localisation on various scales of faces by taking advantages of joint extra-supervised and self-supervised multi-task learning.
One important contribution of this work is that they manually annotate five facial landmarks on the WIDER FACE dataset and observe significant improvement in hard face detection with the assistance of
this extra supervision signal. Another contribution is that they added a self-supervised mesh decoder branch for predicting a pixel-wise 3D shape face information in parallel with the existing supervised branches.
They were able to achieve state of the art performance on several face detection benchmarks.
The overall architecture of RetinaFace model is shown in Fig \ref{fig:retina}.

In \cite{zhu2020tinaface}, Zhu et al. proposed TinaFace, a simple and strong baseline for face detection, which uses ResNet-50 as the feature extraction part, and 6 level Feature Pyramid Network to extract the multi-scale features of input image, followed by an  Inception block to enhance the receptive field. One main purpose of this work was to show that there is no gap between face detection and generic object detection.

In \cite{najibi2019fa}, Najibi et al. proposed a novel approach for generating region proposals for performing face detection, called "Floating Anchor Region Proposal Network (FA-RPN)".
Instead of classifying
anchor boxes using pixel-level features in the convolutional feature map, they generate region proposals using a pooling-based approach.
%
%
The overall architecture of FA-RPN is shown in Fig \ref{fig:fanet}.
\begin{figure}[h]
\centering
\includegraphics[width=0.99\linewidth]{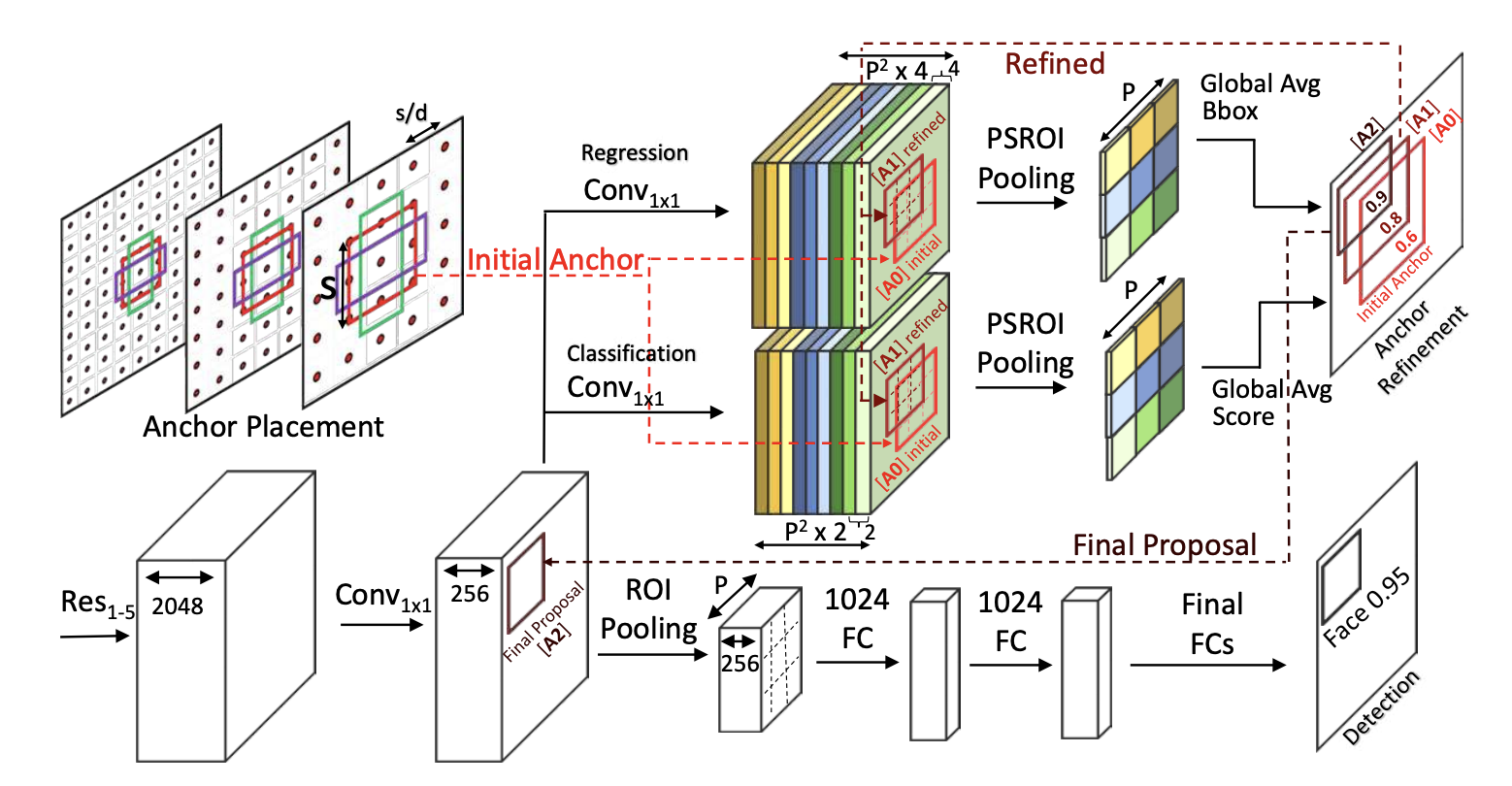}
\caption{The architecture of FA-RPN framework for face detection. Courtesy of \cite{najibi2019fa}}
\label{fig:fanet}
\end{figure}

There are also other works in this category, including robust and high performance face detector \cite{zhang2019robust}, Improved Selective Refinement Network for Face Detection \cite{zhang2019improved}, "Fast cascade face detection with pyramid network" \cite{zeng2019fast}, "Proposal pyramid networks for fast face detection" \cite{zeng2019proposal}, etc.

\subsection{Other Models}
In this section, we cover approaches which do not fall into any of the 
categories mentioned above, or whoaw contribution is not in the modeling part but other factors (such as model optimization).

\begin{figure*}[h]
\centering
\includegraphics[width=0.8\linewidth]{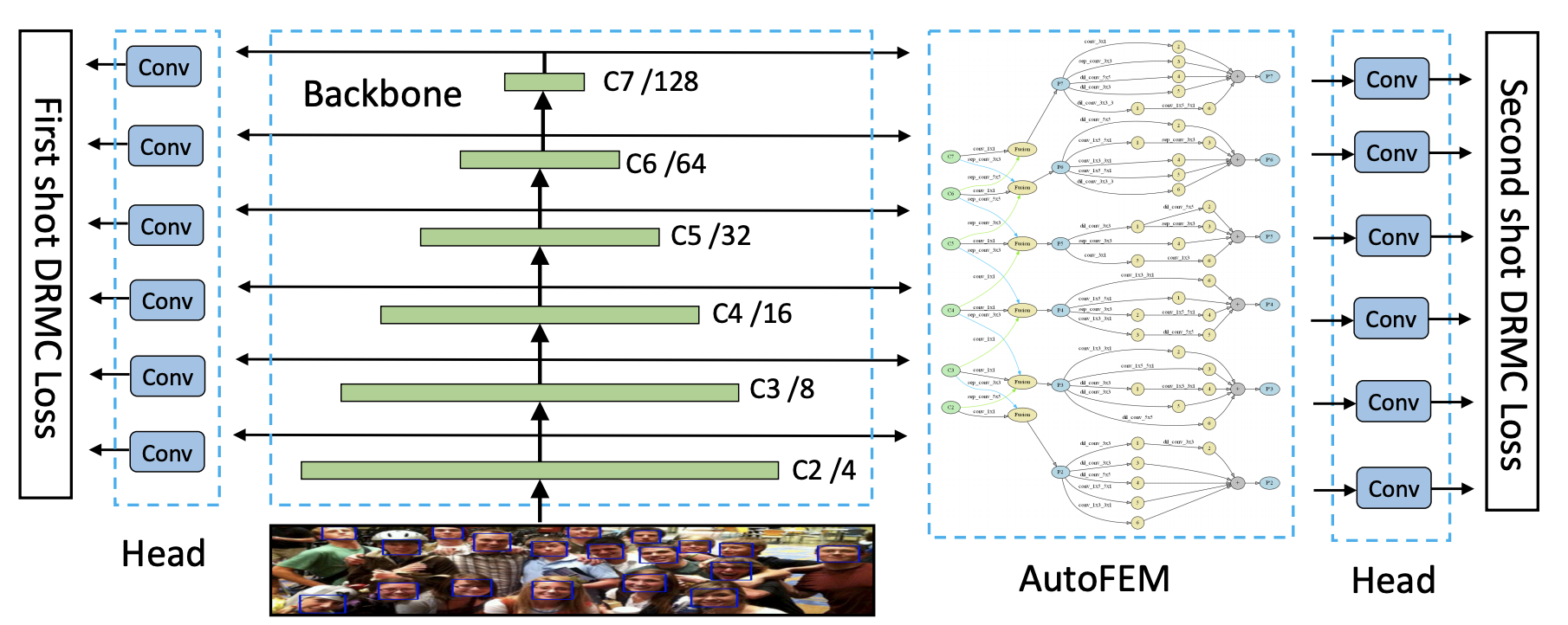}
\caption{The high-level idea of the AFSD framework for face detection. Courtesy of \cite{zhang2020asfd}}
\label{fig:ASFD}
\end{figure*}

In \cite{zhang2020asfd}, Zhang et al. developed  a novel Automatic and Scalable Face Detector (ASFD), which is based on a combination of neural architecture search techniques as well as a new loss design. They proposed an automatic feature enhance module named Auto-FEM by improved differential architecture search, which allows efficient multi-scale feature
fusion and context enhancement. 
They then used Distance-based Regression and Margin-based Classification (DRMC) multi-task loss to predict accurate bounding boxes and learn highly discriminative deep features.
The overall architecture of the proposed AFSD model is shown in Fig \ref{fig:ASFD}.

\begin{figure*}[h]
\centering
\includegraphics[width=0.8\linewidth]{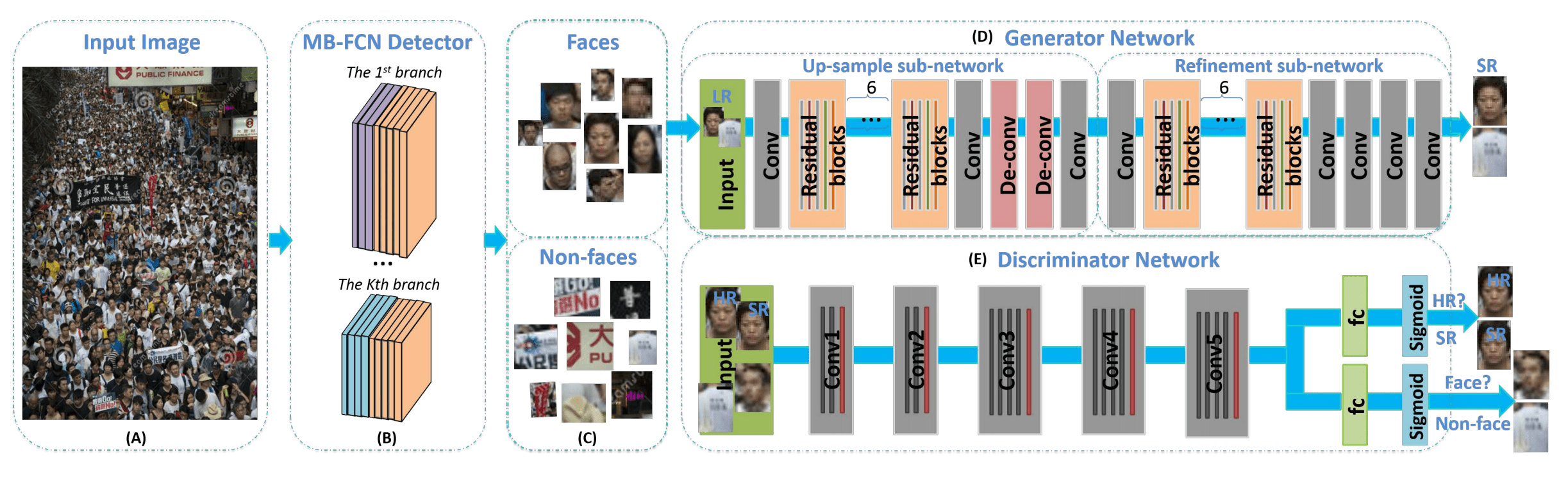}
\caption{The architecture of the proposed tiny face detector based on GANs. Courtesy of \cite{bai2018finding}}
\label{fig:facedet_gan}
\end{figure*}

In \cite{bai2018finding}, Bai et al. proposed a framework for finding tiny faces with a generative adversarial network, which addresses the problem of super-resolving and refining jointly.
Toward this goal, they developed an algorithm to directly generate a clear high-resolution face from a blurry small one by adopting a generative adversarial network (GAN).
The overall architecture of this model is shown in Fig \ref{fig:facedet_gan}.

In \cite{yoo2019extd}, Yoo et al. proposed  a new multi-scale face detector model containing a
tiny number of parameters, called EXTD.
While existing multi-scale face detectors extract feature maps with different scales from a single backbone network, their method generates the feature maps by iteratively reusing a shared lightweight and shallow backbone network.
This iterative backbone sharing strategy significantly reduces the number of parameters, and also provides the abstract image semantics captured from the higher stage of the network layers to the lower-level feature map.

\begin{figure}[h]
\centering
\includegraphics[width=0.99\linewidth]{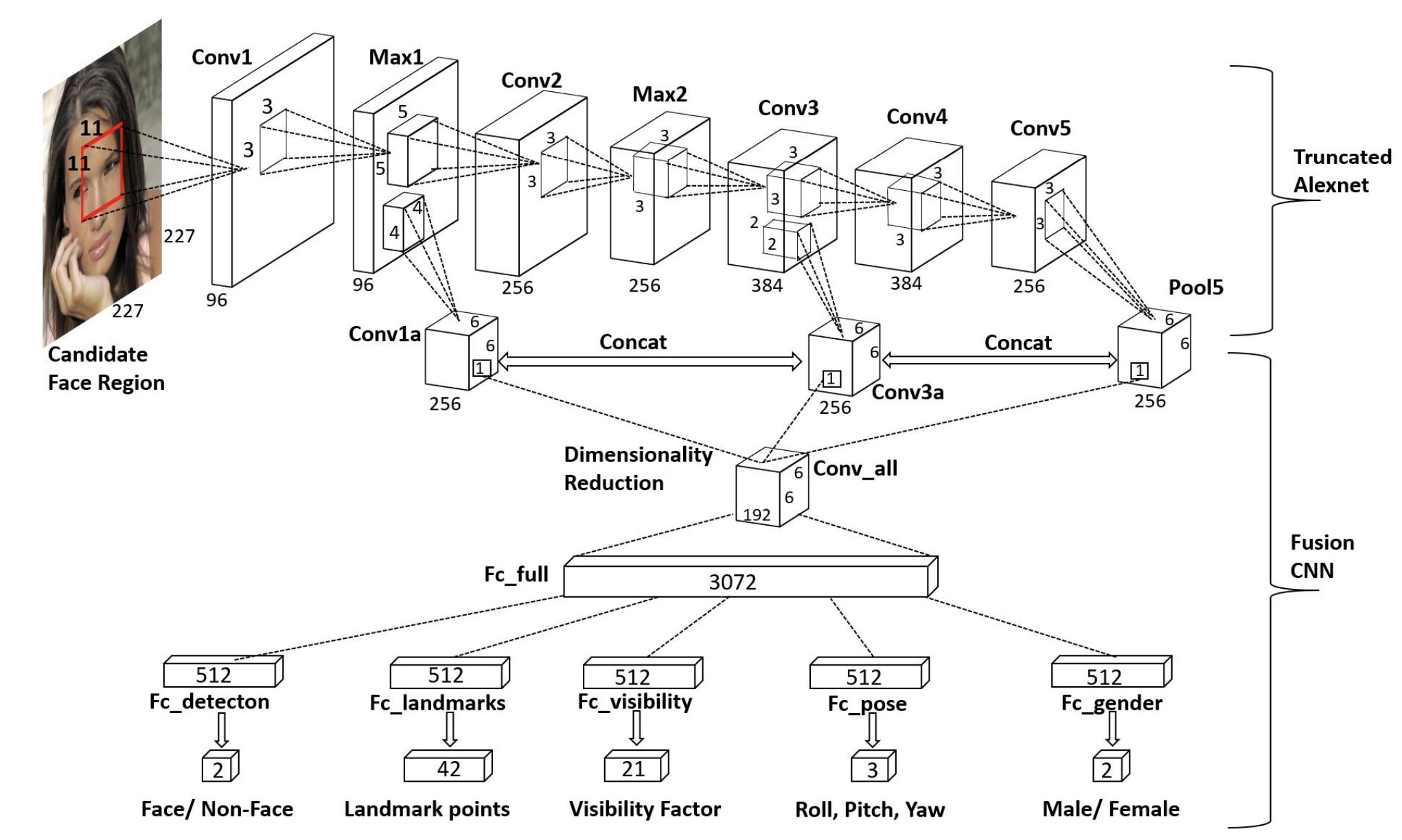}
\caption{The architecture of the HyperFace Model. Courtesy of \cite{ranjan2017hyperface}}
\label{fig:hyper}
\end{figure}

In \cite{ranjan2017hyperface}, Ranjan et al. proposed a framework called HyperFace, for simultaneous face detection, landmarks localization, pose estimation and gender recognition
using deep convolutional neural networks (CNN). The proposed method 
fuses the intermediate layers of a deep CNN using a separate CNN followed by a multi-task learning algorithm that operates on the fused features. It exploits the synergy
among the tasks which boosts up their individual performances.
The architecture of this framework is shown in Fig \ref{fig:hyper}.

\begin{figure}[h]
\centering
\includegraphics[width=0.99\linewidth]{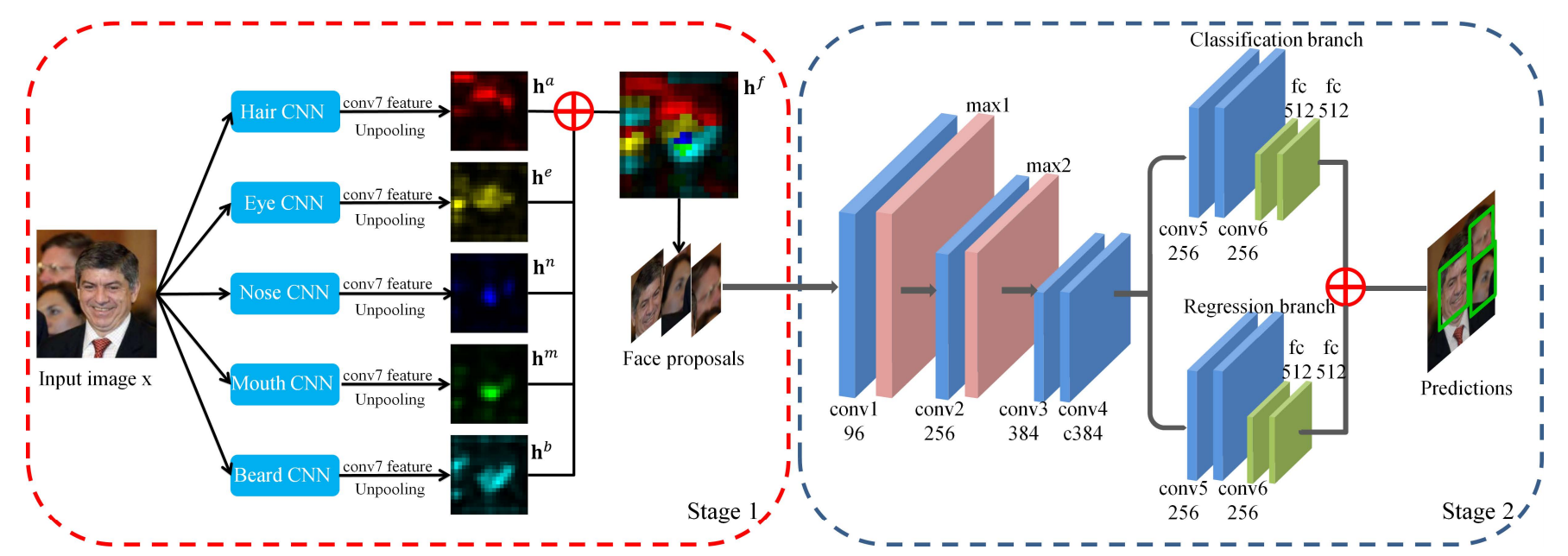}
\caption{The architecture of the Faceness-Net framework. Courtesy of \cite{yang2017faceness}}
\label{fig:faceness}
\end{figure}

In \cite{yang2017faceness}, Yang et al. proposed Faceness-Net, a deep convolutional neural network (CNN) for face detection leveraging on facial attributes based supervision.
They observed a phenomenon that part detectors emerge within CNN trained to classify attributes from uncropped face images, without any explicit part supervision. The observation motivates a new method for finding faces through scoring facial parts responses by their spatial structure and arrangement. 
Fig \ref{fig:faceness} illustrates the architecture of the proposed Faceness-Net.
The first stage of Faceness-Net applies attribute-aware networks to generate response maps of different facial parts. The maps are subsequently employed to produce face proposals. The second stage of Faceness-Net refines candidate windows
generated from first stage using a multi-task convolutional neural network (CNN), where face classification and bounding box regression are jointly optimized. 

\begin{figure}[h]
\centering
\includegraphics[width=0.99\linewidth]{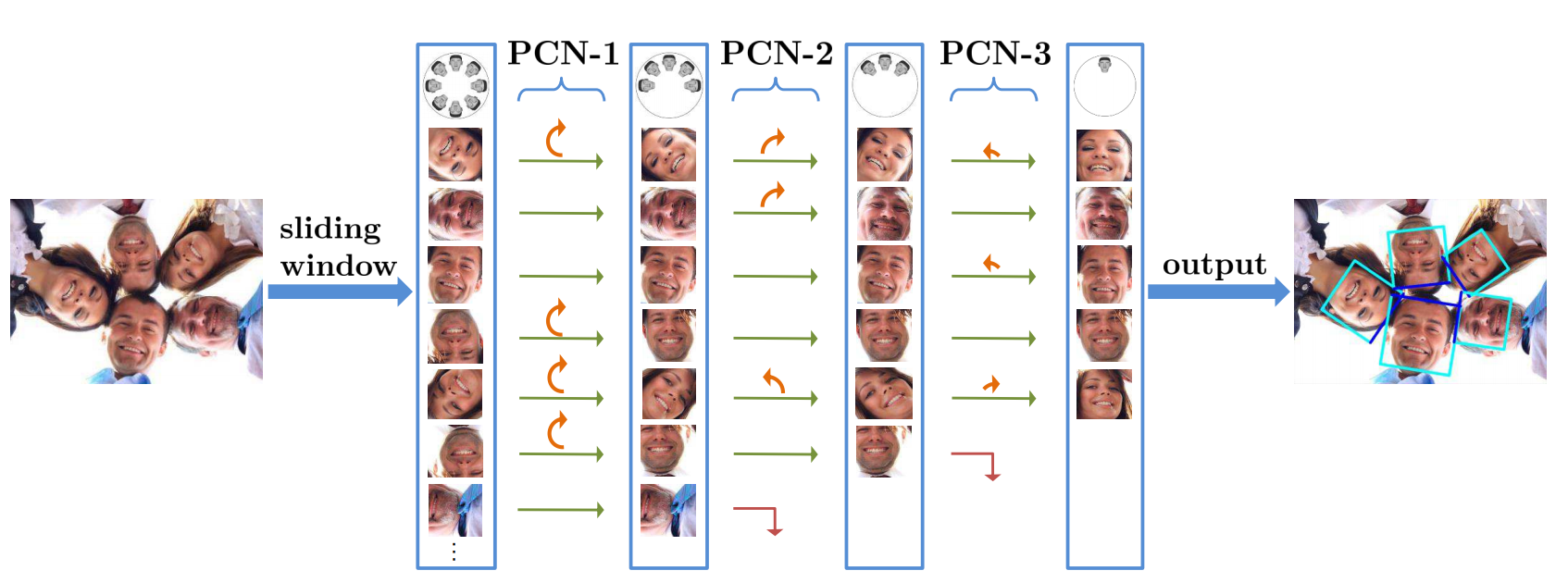}
\caption{The architecture of the progressive calibration networks (PCN) for face detection. The PCN progressively calibrates the RIP orientation of each face candidate to upright for better distinguishing faces from non-faces. Specifically, PCN-1 first
identifies face candidates and calibrates those facing down to facing up.
Then the rotated face candidates are further distinguished and calibrated to an upright range of $[-45,45]$ in PCN-2, shrinking the RIP ranges by half again. Finally, PCN-3 makes the accurate final decision for each face candidate to determine whether it is a face and predict the precise RIP angle. Courtesy of \cite{shi2018real}}
\label{fig:PCN}
\end{figure}

In \cite{shi2018real}, Shi et al. proposed a real-time rotation-invariant face detection called "Progressive Calibration Networks". 
PCN consists of three
stages, each of which not only distinguishes the faces from non-faces, but also calibrates the rotation-in-plane orientation of each
face candidate to upright progressively. By dividing the calibration process into several progressive steps and only predicting coarse orientations in early stages, PCN can
achieve precise and fast calibration.
Figure \ref{fig:PCN} illustrates the high-level architecture of the PCN framework.

In \cite{cheng2019segmentation}, Cheng et al. proposed an anchor-free and non-maxima-supression-free object detection model, called weakly supervised multi-modal annotation segmentation (WSMA-Seg), which utilizes segmentation models to achieve an accurate and robust object detection without NMS.
In addition to this, they proposed a multi-scale pooling segmentation (MSP-Seg) as the underlying segmentation model of WSMA-Seg to achieve a more accurate segmentation and to enhance the detection accuracy of WSMA-Seg.
Through experimental results on multiple datasets they show that the proposed WSMA-Seg approach achieves promising results on several detection benchmarks, including widerFace.

In \cite{saha2020rnnpool}, Saha et a. introduced RNNPool, a novel pooling operator based on Recurrent Neural Networks (RNNs), that efficiently aggregates features over large patches of an image and rapidly downsamples activation maps, which they claimed to be more suitable for computer vision tasks, such as face detection.
Empirical evaluation indicates that an RNNPool layer can effectively replace multiple blocks in a variety of architectures such as MobileNets, DenseNet when applied to standard vision tasks like  face detection.

Some of the other popular models for face detection include: "A Light and Fast Face Detector for Edge Devices" \cite{he2019lffd}, "Accurate face detection for high performance" \cite{zhang2019accurate}, UnitBox \cite{yu2016unitbox}, "HAMBox: Delving Into Mining High-Quality Anchors on Face Detection" \cite{liu2020hambox},  "Joint face detection and facial motion re-targeting for multiple faces" \cite{chaudhuri2019joint}, "Hierarchical attention for part-aware face detection" \cite{wu2019hierarchical}, "Group Sampling for Scale Invariant Face Detection" \cite{ming2019group}, "Dafe-fd: Density aware feature enrichment for face detection" \cite{sindagi2019dafe}, "Triple loss for hard face detection" \cite{fang2020triple}, "BlazeFace: Sub-millisecond Neural Face Detection on Mobile GPUs" \cite{bazarevsky2019blazeface}, and Face Detection with End-to-End Integration of a ConvNet and a 3D Model \cite{li2016face}.
%
%

It is worth providing some high-level comparison between different categories of face detection models. Cascade-CNN based models are designed for efficient high performance face detectors which may be suitable for deployment in edge devices, or camera capture applications. However, these models is hard to bring the state of the art accuracy on challenging cases with low image quality, non-standard poses and lightings. On the other hand, the general object detector-based pipelines such as R-CNN can provide much better accuracy due to the more capacity and learning power of the model architectures used. The single-shot or anchor-free detectors such as SSD, FCOS, etc. provides a good tradeoff between the accuracy and efficiency. For backbone architectures, feature pyramid-based architectures have been shown to be outperforming standard CNN backbones for object detection without much degradation in efficiency. Furthermore, recently transformer-based architectures have shown state of the art results on object detection, and provides a potential promising avenue  for future extensions of existing models for boosting the performance of face detection to the next level. Finally other approaches tackling specific challenges in face detection such as GAN-based approaches, or PCN detector specifically designed for better handling of rotated faces, are complementary to existing mainstream approaches.

\section{Face Detection Benchmark Datasets}
\label{sec:datasets}
Several datasets have been proposed for face detection over the past decades. 
Here, we provide an overview of the most widely used datasets in the recent literature.
Important factors to consider in comparing datasets include: 
number of images and faces, 
size range of faces,
amount of meta-data specified for each face,
and
range of face/image quality conditions represented.

\subsection{FDDB}
\begin{figure}[h]
\centering
\includegraphics[width=0.9\linewidth]{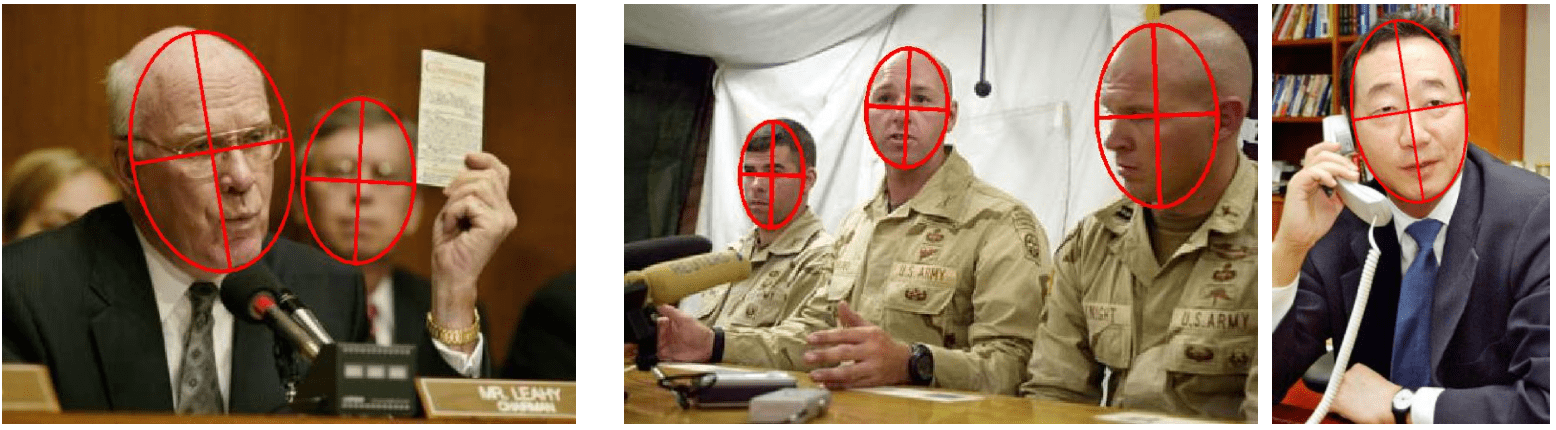}
\caption{Three sample images from FDDB dataset, and the ground-truth face regions. Courtesy of \cite{jain2010fddb}}
\label{fig:fddb}
\end{figure}

Face Detection Data Set (also known as FDDB) contains the annotations for 5171 faces in 2845 images taken from the Labelled Faces in the Wild data set \cite{jain2010fddb}.
FDDB contains a wide range of difficult elements, including occlusions, difficult poses, and low resolution and out-of-focus faces.
The specifications of face regions are provided as elliptical regions.
They proposed an evaluations metric based on ROC,
coarse and precise score metrics to match between prediction and ground-truth.
Three sample images from this dataset are shown in Fig \ref{fig:fddb}.

\subsection{WIDER FACE (2016)}
\begin{figure}[h]
\centering
\includegraphics[width=0.9\linewidth]{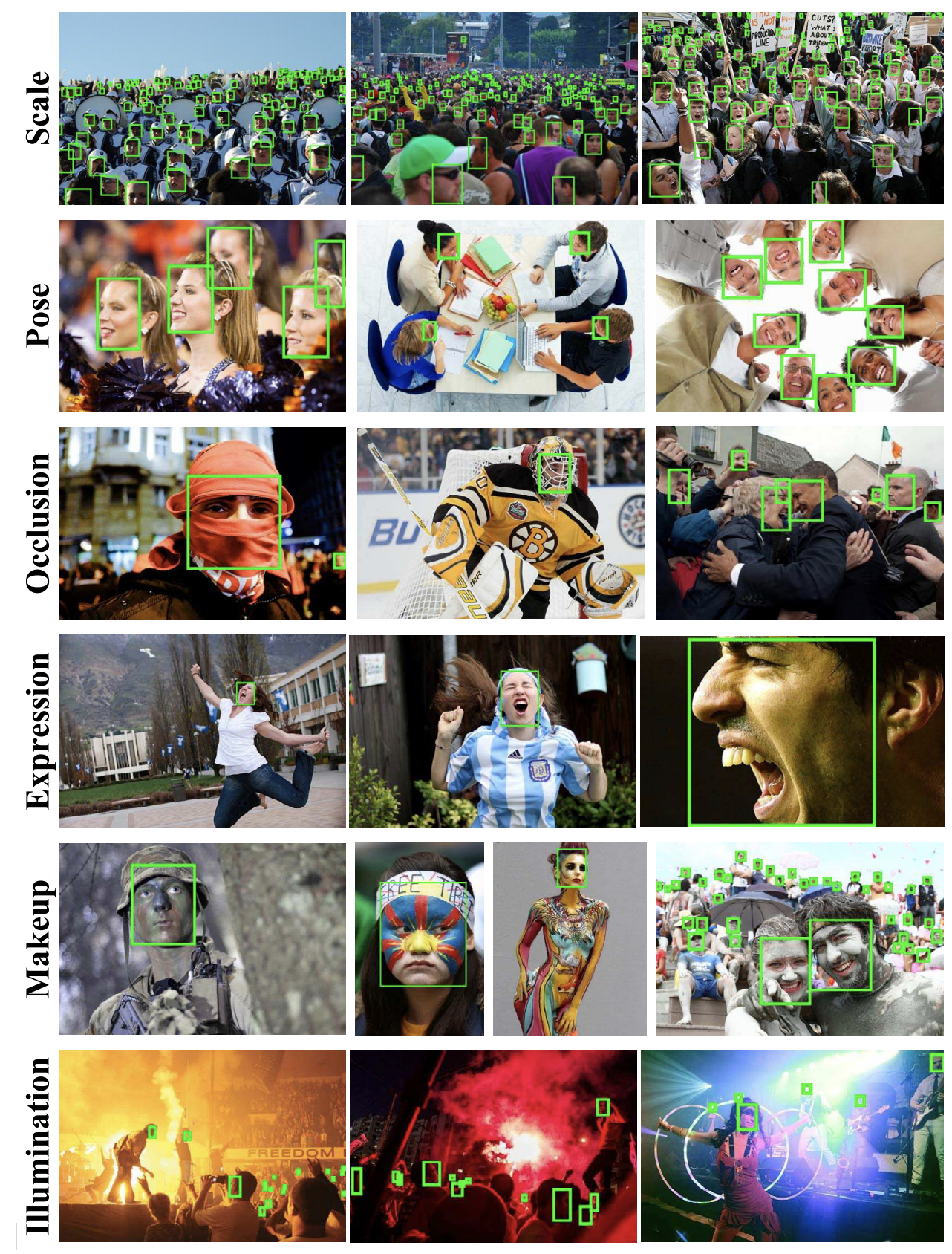}
\caption{Some of the sample images from Wider-Face dataset, and the ground-truth face bounding boxes. Courtesy of \cite{yang2016wider}}
\label{fig:wider}
\end{figure}

The Wider Face dataset \cite{yang2016wider} 
was introduced after FDDB and 
contains 32,203 images and 393,703 labeled faces with a high degree of variability in scale, pose and occlusion, making it a very challenging dataset. 
The images are selected from the publicly available Wider dataset.
The Wider Face dataset is organized based on 61 event classes. For each event class, they randomly selected 40\%, 10\%, 50\% data as training, validation and testing sets accordingly. They adopted the same evaluation metric employed in the PASCAL VOC dataset. Similar to some of the other datasets, they do not release bounding box ground truth for the test images. Users are required to submit final prediction files, which they shall proceed to evaluate.
Fig \ref{fig:wider} illustrates some of the sample images from this dataset. As we can see there is a high degree of variability in scale, pose, occlusion, expression, appearance and illumination.


\subsection{PASCAL Face}
PASCAL face dataset contains 1,335 labeled faces in 851 images with large face appearance and pose variations \cite{yan2014face}. It is collected from PASCAL person layout test subset.
This dataset is small compared to other face detection datasets.

\subsection{MALF}
\begin{figure}[h]
\centering
\includegraphics[width=0.99\linewidth]{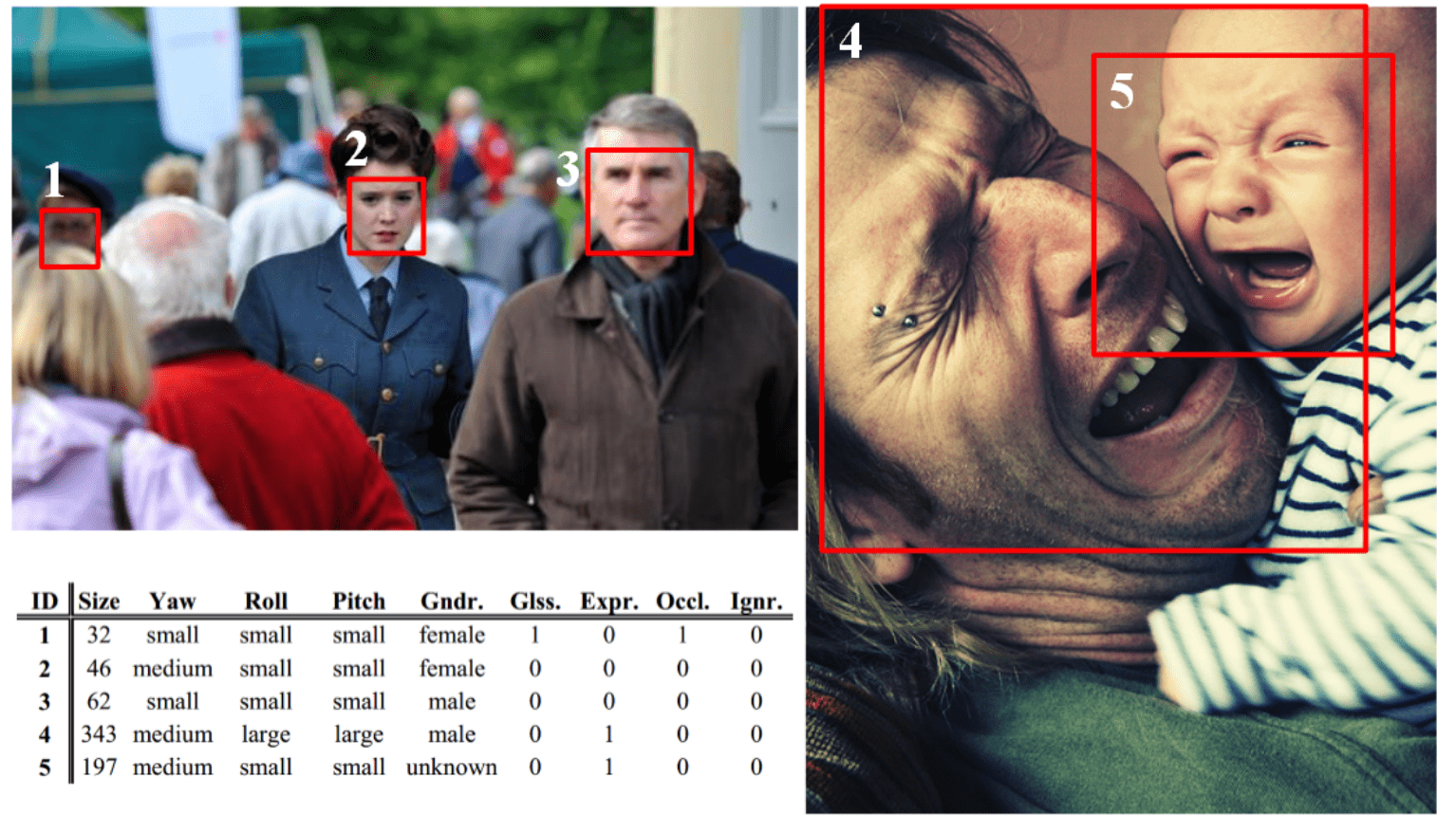}
\caption{Two sample images and their annotations from MALF dataset. Courtesy of \cite{yang2015fine}}
\label{fig:malf}
\end{figure}

Multi-Attribute Labelled Faces (MALF) is the first face detection dataset that supports fine-gained evaluation. MALF consists of 5,250 images and 11,931 faces \cite{yang2015fine}.
Besides bounding box annotations, each face contains other annotations such as: pose deformation level of yaw, pitch and roll (small, medium, large); other facial attributes: gender(female, male, unknown), is-Wearing-Glasses, is-occluded and is-Exaggerated-Expression.
Fig \ref{fig:malf} shows some of the sample images and annotations from MALF.

\subsection{UFDD (2018)}
\begin{figure}[h]
\centering
\includegraphics[width=0.99\linewidth]{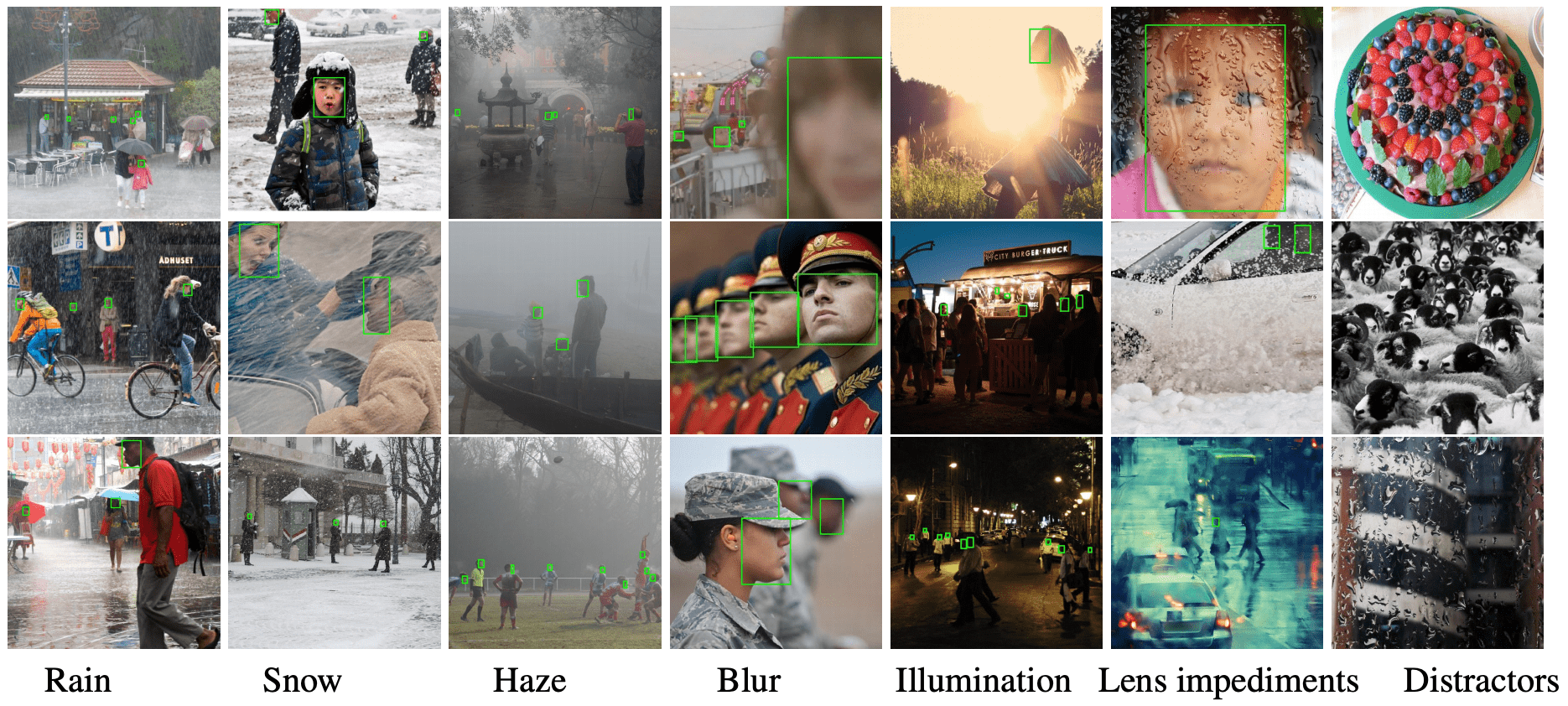}
\caption{Sample images from UFDD dataset. Courtesy of \cite{nada2018pushing}}
\label{fig:ufdd}
\end{figure}

Unconstrained Face Detection Dataset (UFDD) \cite{nada2018pushing} contains a total of 6,425 images with 10,897 face-annotations.
It involves key degradations or conditions including: rain, snow, haze, lens impediments, blur, illumination variations, and distractors.
Fig \ref{fig:ufdd} shows some sample images from this dataset with different variations and the corresponding annotations.


\subsection{VGGFace2 Dataset}
VGGFace2 dataset contains 3.31 million
images of 9131 subjects, with an average of 362.6 images for each subject \cite{cao2018vggface2}. Images are downloaded from Google Image Search and have large variations in pose, age, illumination,
ethnicity and profession (e.g. actors, athletes, politicians).
Faces in this dataset are detected using the model provided by \cite{zhang2016joint}.

\section{Experimental Performance}
\label{sec:performance}
In this section, we first introduce some of the popular metrics used to evaluate the performance of face detection models.
Then we summarize the quantitative performance of the more promising DL-based face detection models on popular datasets.

\subsection{Evaluation Metrics}
There are a few metrics which are widely used for evaluating the performance of face detection models, including: Average Precision (AP), Precision-Recall (PR) Curve, and receiver operating characteristic (ROC) curve.
We give brief definitions of these metrics below.

\subsubsection{Precision-Recall Curve}
Before getting into the description of precision-recall curve, we need to give the mathematical definition of precision and recall metrics. These metrics are defined as Eq \ref{prec_rec}:
\begin{equation}
\text{Precision} =  \frac{\text{TP}}{\text{TP}+\text{FP}}; \qquad
\ \ \text{Recall} =  \frac{\text{TP}}{\text{TP}+\text{FN}},
\label{prec_rec}
\end{equation}

Now assuming that the model output is a score in [0,1], we can compare the output with a threshold to get the class label. Therefore, for each threshold value we can find the corresponding precision and recall values. 
\textbf{Precision-Recall curve} shows the precision as a function of recall, for all different threshold values.

\subsubsection{Average Precision (AP)}
As mentioned above, a model's precision refers to precision at a particular decision threshold.
Average precision calculates the average precision at all such possible thresholds, which is also similar to the area under the precision-recall curve. It is a useful metric to compare how well model's predictions are, without considering any specific decision threshold.

\subsubsection{ROC Curve}
The receiver operating characteristic (ROC) curve is plot which shows the performance of a model as a function of its cut-off threshold (similar to precision-recall curve). It essentially shows the true positive rate (TPR) against the false positive rate (FPR) for various threshold values.
In general, the lower the cut-off threshold on positive class, the more samples predicted as positive class, i.e. higher true positive rate (recall) and also higher false positive rate (corresponding to the right side of this curve). Therefore, there is a trade-off between how high the recall could be versus how much we want to bound the error (FPR).

\subsection{Quantitative Performance of DL-Based Face Detection Models}
\label{sec:quant_result}

In this section we tabulate the performance of several of the previously discussed algorithms on popular face detection benchmarks. 
Table \ref{table_wider} provides the performance of some of the prominent works on Wider-Face dataset. It is worth mentioning that there are three versions available for the Wider-Face test set. We report the performance of each moodel on all three versions, when they are available.
As we can see, there has been a huge progress on the performance of deep models on this dataset. 
Tables \ref{table_fddb} and \ref{table_pascal} provide the performance of some of the deep learning based face detection models on FDDB, and PASCAL Face datasets, respectively. As it can be seen, the performances on these two datasets are typically higher than those on Wider-Face.
 
\begin{table}
\centering
\caption{The performance of face detection models on different versions of Wider-Face dataset.}
\label{table_wider}
\begin{tabular}{lccc}
\toprule
Method & Easy & Medium  &  Hard \\
\midrule
Faceness \cite{yang2017faceness}  & 71.3  & 63.4  & 34.5 \\
Multiscale Cascade CNN \cite{yang2016wider}  & 69.1  & 66.4  & 42.4 \\
CMS-RCNN \cite{zhu2017cms}  & 90.2  & 87.4  & 64.3 \\
LFFD \cite{he2019lffd}  & 89.6 & 86.5  & 77.0 \\
img2pose \cite{albiero2020img2pose}  & 90.0 & 89.1  & 83.9 \\
S3FD \cite{zhang2017s3fd}  & 92.8 & 91.3  & 84.4 \\ 
EXTD \cite{yoo2019extd}  & 91.2  & 90.3  & 85.0 \\ 
FACE R-FCN \cite{wang2017detecting}  & 94.3  & 93.1  & 87.6 \\ 
SRN \cite{chi2019selective}  & 95.9  & 94.8  & 89.6 \\ 
FDNet \cite{zhang2018face}  & 95.0  & 93.9  & 89.6 \\
DSFD \cite{li2019dsfd}  & 96.0  & 95.3  & 90.0 \\ 
PyramidBox \cite{tang2018pyramidbox}  & 95.6  & 94.6  & 90.0 \\ 
AInnoFace \cite{zhang2019accurate}  & 96.5  & 95.7  & 91.2 \\ 
RetinaFace \cite{deng2019retinaface}  & -  & -  & 91.4 \\
TinaFace \cite{zhu2020tinaface}  & -  & -  & 92.4 \\
\bottomrule
\end{tabular}
\end{table}


\begin{table}
\centering
\caption{The performance of face detection models on the FDDB dataset.}
\label{table_fddb}
\begin{tabular}{lc}
\toprule
Method   &  AP \\
\midrule
CascadeCNN \cite{li2015convolutional}  & 85.7 \\
Joint-Cascade \cite{qin2016joint}  & 86.3 \\
HyperFace \cite{ranjan2017hyperface}  & 90.1 \\
Faceness \cite{yang2017faceness}  & 90.3 \\
DP2MFD \cite{chi2019selective}  & 90.3 \\
UnitBox \cite{yu2016unitbox}  & 95.1 \\
FaceBoxes \cite{zhang2017faceboxes}  & 96.0 \\
Faster R-CNN \cite{ren2015faster}  & 96.1 \\
DPSSD \cite{ranjan2019fast}  & 96.1 \\
LFFD \cite{he2019lffd}  & 97.3 \\
S3FD \cite{zhang2017s3fd}   &  98.3 \\
PyramidBox \cite{tang2018pyramidbox}    &  98.7 \\ 
SRN \cite{chi2019selective}   &  98.8 \\ 
FACE R-FCN \cite{wang2017detecting}    & 99.0 \\ 
DSFD \cite{li2019dsfd}   &  99.1 \\
\bottomrule
\end{tabular}
\end{table}


\begin{table}
\centering
\caption{The performance of face detection models on PASCAL Face dataset.}
\label{table_pascal}
\begin{tabular}{lc}
\toprule
Method   &  AP \\
\midrule
Headhunter   &  89.63 \\
DPM \cite{liao2015fast}  & 90.29 \\
Faceness \cite{yang2017faceness}  & 92.11 \\
STN \cite{chen2016supervised}  & 94.10 \\
HyperFace \cite{ranjan2017hyperface}  & 96.20 \\
FaceBoxes \cite{zhang2017faceboxes}  & 96.30 \\
S3FD \cite{zhang2017s3fd}  & 98.49 \\
Anchor-based \cite{zhang2020robust}  & 99.00 \\
SRN \cite{chi2019selective}  & 99.09 \\
\bottomrule
\end{tabular}
\end{table}

As we can see the performance of the recent models on the FDDB and PASCAL face datasets is higher than 99\%, which means that even with a much more powerful and novel model we will not be able to see a significant quantitative gain over the previous models. Hence, there is a real need for developing of larger and harder datasets of face detection.

\begin{figure}[h]
\centering
\includegraphics[width=0.845\linewidth]{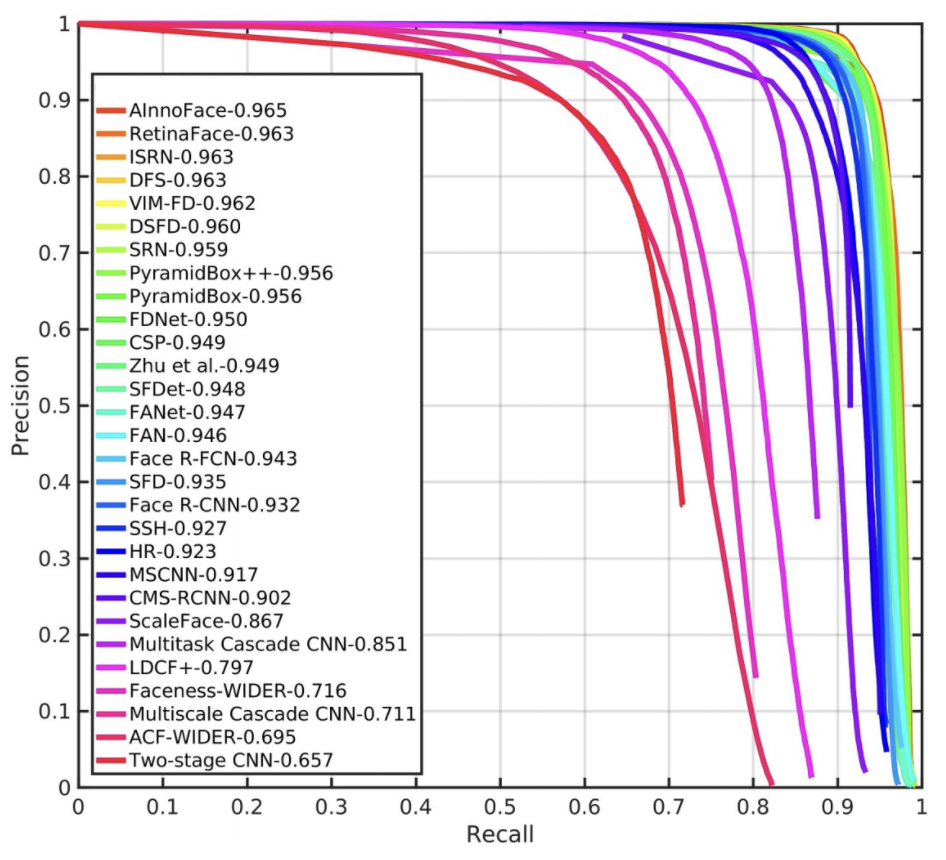}
\includegraphics[width=0.845\linewidth]{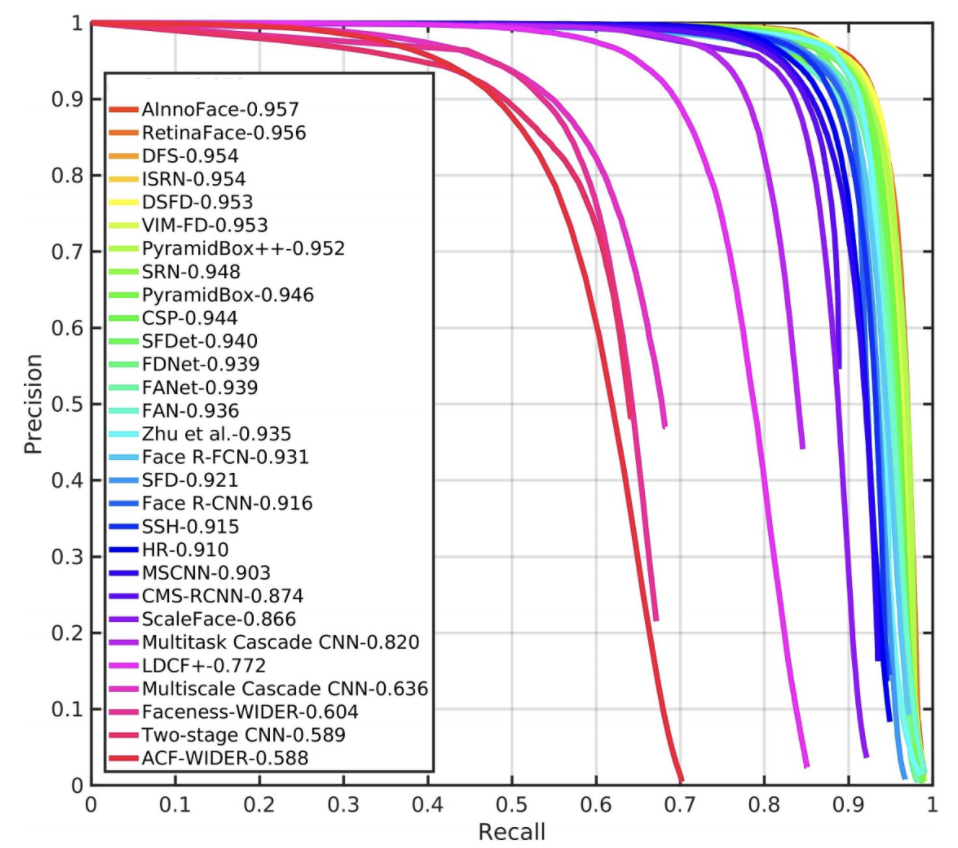}
\includegraphics[width=0.83\linewidth]{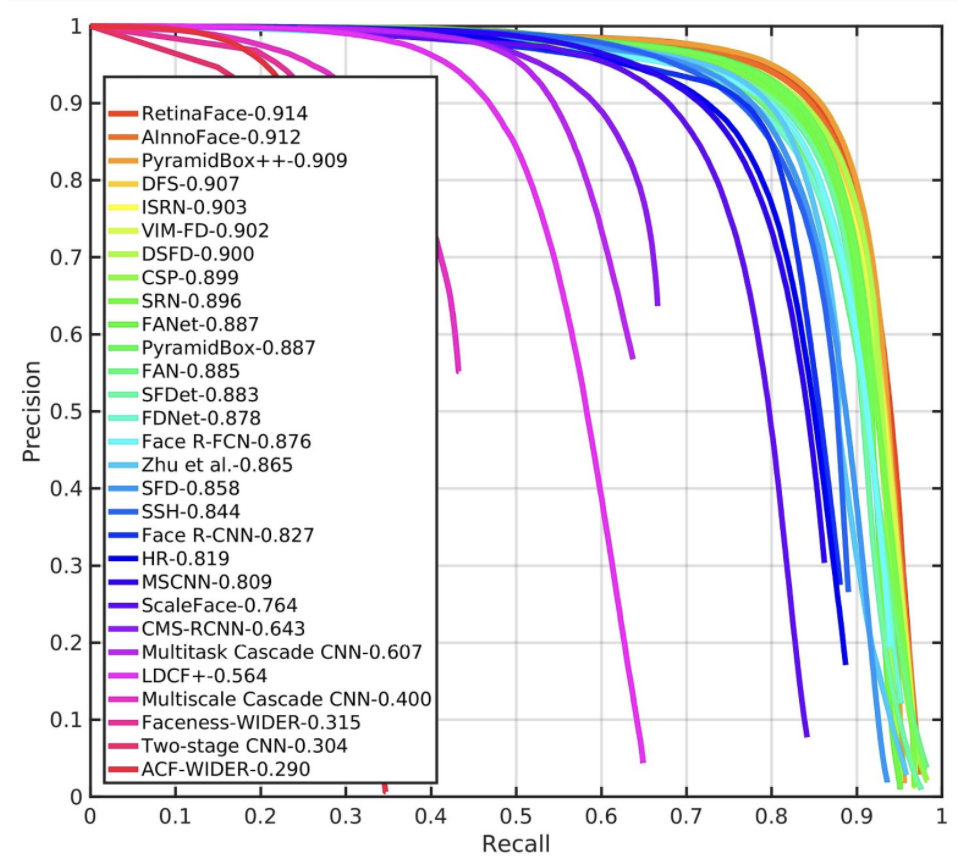}
\caption{The top, middle, and bottom figures present the Precision-Recall curves of promising face-detection models on WiderFace Easy/Medium/Hard test sets, respectively. Courtesy of \cite{zhang2020refineface}.}
\label{fig:roc_wider}
\end{figure}

\begin{figure}[h]
\centering
\includegraphics[width=0.99\linewidth]{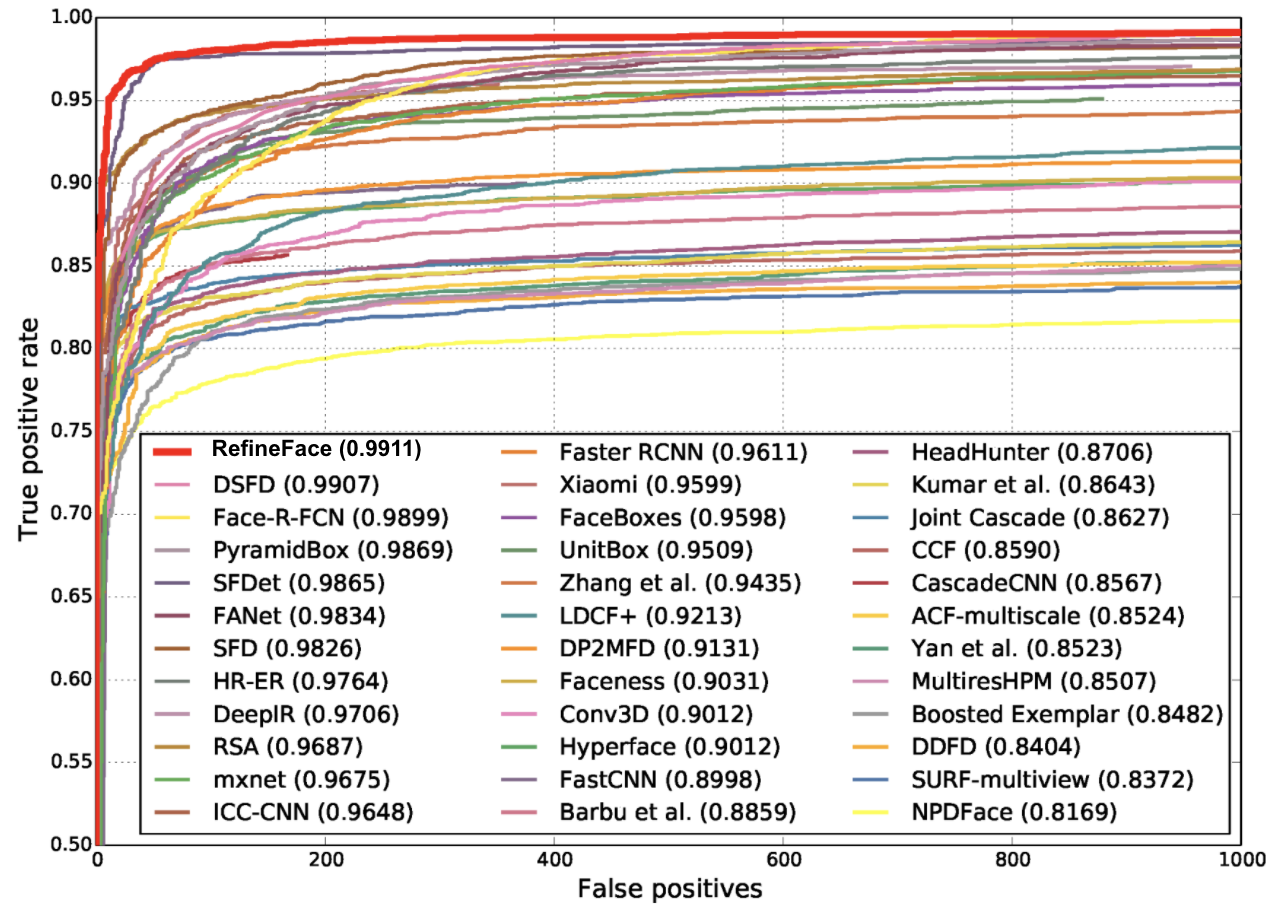}
\caption{The Receiver operating characteristic (ROC) curves of face detection models on FDDB benchmark. Courtesy of \cite{zhang2020refineface}}
\label{fig:roc_fddb}
\end{figure}

In addition to the average precision of the models listed above, we provide the Precision-Recall curves of those models on WiderFace dataset (in Fig \ref{fig:roc_wider}), and their Receiver operating characteristic (ROC) curves on FDDB benchmark (in Fig \ref{fig:roc_fddb}).

\begin{figure*}[h]
\centering
\includegraphics[width=0.8\linewidth]{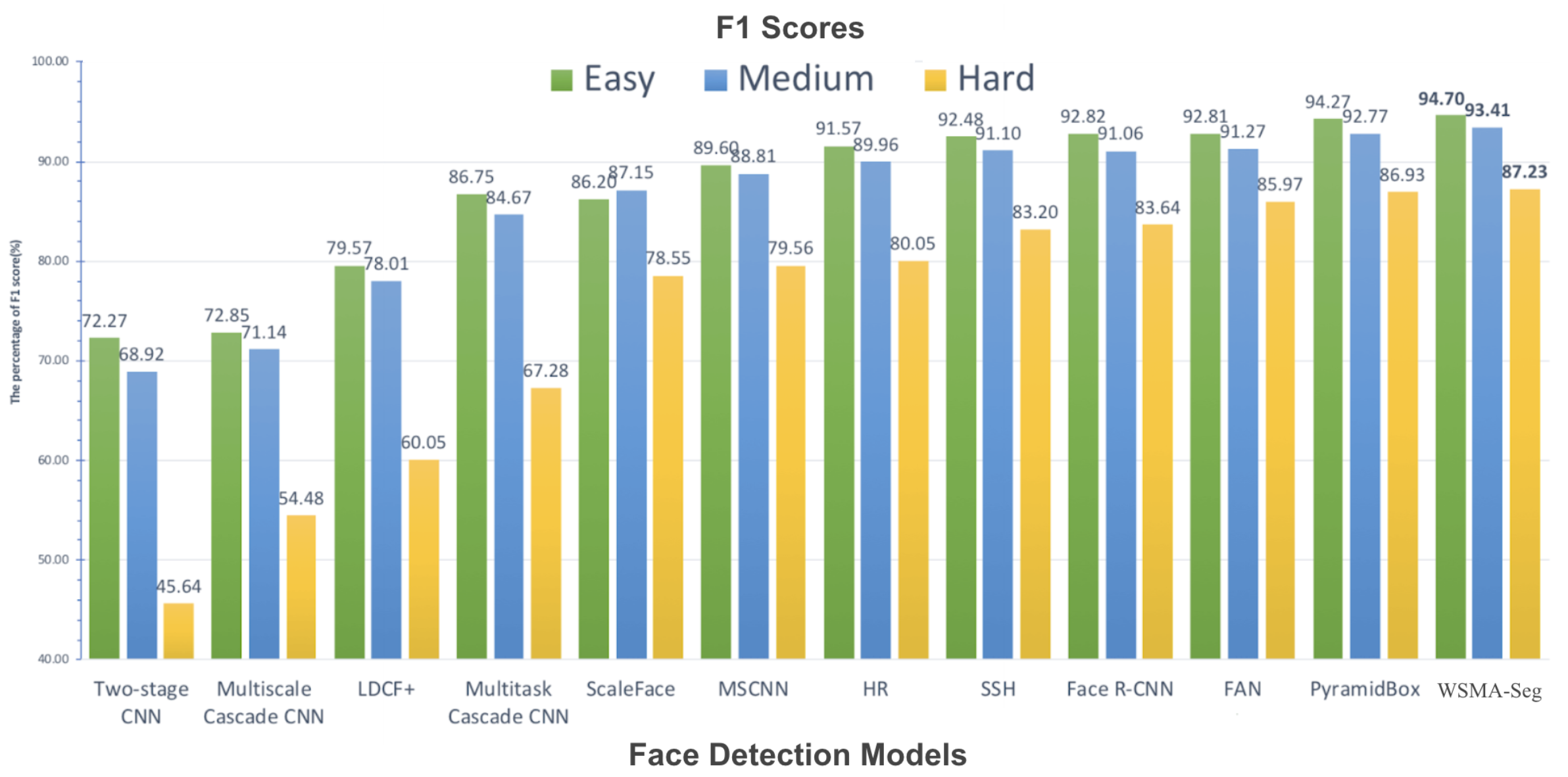}
\caption{The F1 scores of face detections models on the WiderFace benchmark. Courtesy of \cite{cheng2019segmentation}}
\label{fig:f1_scores}
\end{figure*}

We also provide a comparison of some of the prominent face detection models, in terms of F1-score, in Fig \ref{fig:f1_scores}

\section{Challenges and Opportunities}
\label{sec:challenges}
With no doubt, face detection has seen a great improvement over the last few years, thanks to new deep learning based models.
However, several challenges lie ahead. We will next discuss some of the promising research directions that we believe will help in further advancing face detection.

\subsection{Detection Robustness on Tiny Faces}
Several researchers have already focused specific efforts on detecting faces that appear in tiny regions in images.
However, 
there is still much room for improving the accuracy of these models for detecting tiny faces in general, and the more challenging cases of tiny faces with heavy occlusion, or facial disguise.

\subsection{Face Occlusion}
Occlusion is one of the most critical challenges in face detection.
Various factors can lead to occlusion, such as: accessories (Goggles, Cap), medical masks, beard/moustache, facial disguise, or blockage by another person or object. 
Although many of the current face detection models handle occlusion to some extent, it is still very challenging to detect faces with heavy occlusion. 
There are two avenues to improve face detection models with occlusion. First, to develop new large-scale datasets of faces with occlusion. Second, to develop model architectures which are better curated toward detecting faces with occlusion.

\subsection{Accurate Lightweight Models}
Face detection is an important step in many real-world problems and use-cases, and therefore it is becoming deployed in various mobile applications. 
Hence, it is important for these models to be lightweight and memory efficient, in addition to obtaining high accuracy.
This can  be achieved either by investigating novel lightweight model architectures in the first place, or by applying model compression and neural architecture search techniques to already-developed models.

\subsection{Few-Shot Face Detection}
Most of the current face detection models are trained on very large-scale datasets of annotated faces, and there have not been 
many deep-learning based works for face detection from few labeled samples. This could be specially useful for new use-cases for which a large-scale dataset is not available, such as animation face detection, or detecting all faces in the crowd of images (for which there usually exist more than 100 faces per image). 

\subsection{Interpretable Deep Models}
While DL-based models have achieved promising performance on
challenging face detection benchmarks, there remain open questions about these models. For example, what exactly are deep models learning? What is a minimal neural architecture that can achieve a certain  accuracy on a given dataset? 
How should we interpret the features learned by these models?
Although there are some techniques available to visualize the learned convolutional kernels of these models, a comprehensive study of the underlying dynamics of these models is lacking. A better understanding of the theoretical aspects of these models can enable us to develop better models curated for face detection.

\subsection{Face Detection Bias Reduction}
The topic of ``bias'', or accuracy disparity across demographic groups, is a hot issue currently.
Researchers are actively engaged on this topic for face recognition \cite{Albiero_2020,Cavazos_2021,Krishnapriya_2020,dhar2020adversarial} and for analytics such as gender-from-face \cite{Drozdowski_2020}.
However, we know of no work that has focused specifically on possible bias in face detection.
Experimental study of this topic presents some non-trivial challenges.
The meta-data for an experimental dataset should not be created using a face detection algorithm, or the dataset may inherit any bias in the algorithm.
Also, factors that are likely to affect accuracy in general should be balanced across demographic groups in the dataset, so that observed disparities can be inferred to be due to demographic factors.

\section{Summary and Conclusions}
\label{sec:conclusions}
We have surveyed recent face detection methods based on deep learning models, which have achieved promising results on various face detection benchmarks and enabled the usage of these models on real-world applications. 
We categorized these models into several architectural groups: Cascaded CNN models, R-CNN based models, SSD-based models, and FPN-based models, and identified their main technical contributions.
We also provided an overview of some of the popular face detection benchmarks, such as Wider-Face, FDDB, and PASCAL Face.
We then summarized the quantitative performance of these
models on these popular benchmarks. 
Finally, we discussed some of the open challenges and promising directions for deep-learning-based face detection in the coming years.

\section*{Acknowledgments}
The authors would like to thank Aleksei Stoliar for his comments and suggestions regarding this work.


\bibliographystyle{IEEEtran}
\bibliography{pami20}



\begin{IEEEbiography}[{\includegraphics[width=0.95in,height=1.25in,clip,keepaspectratio]{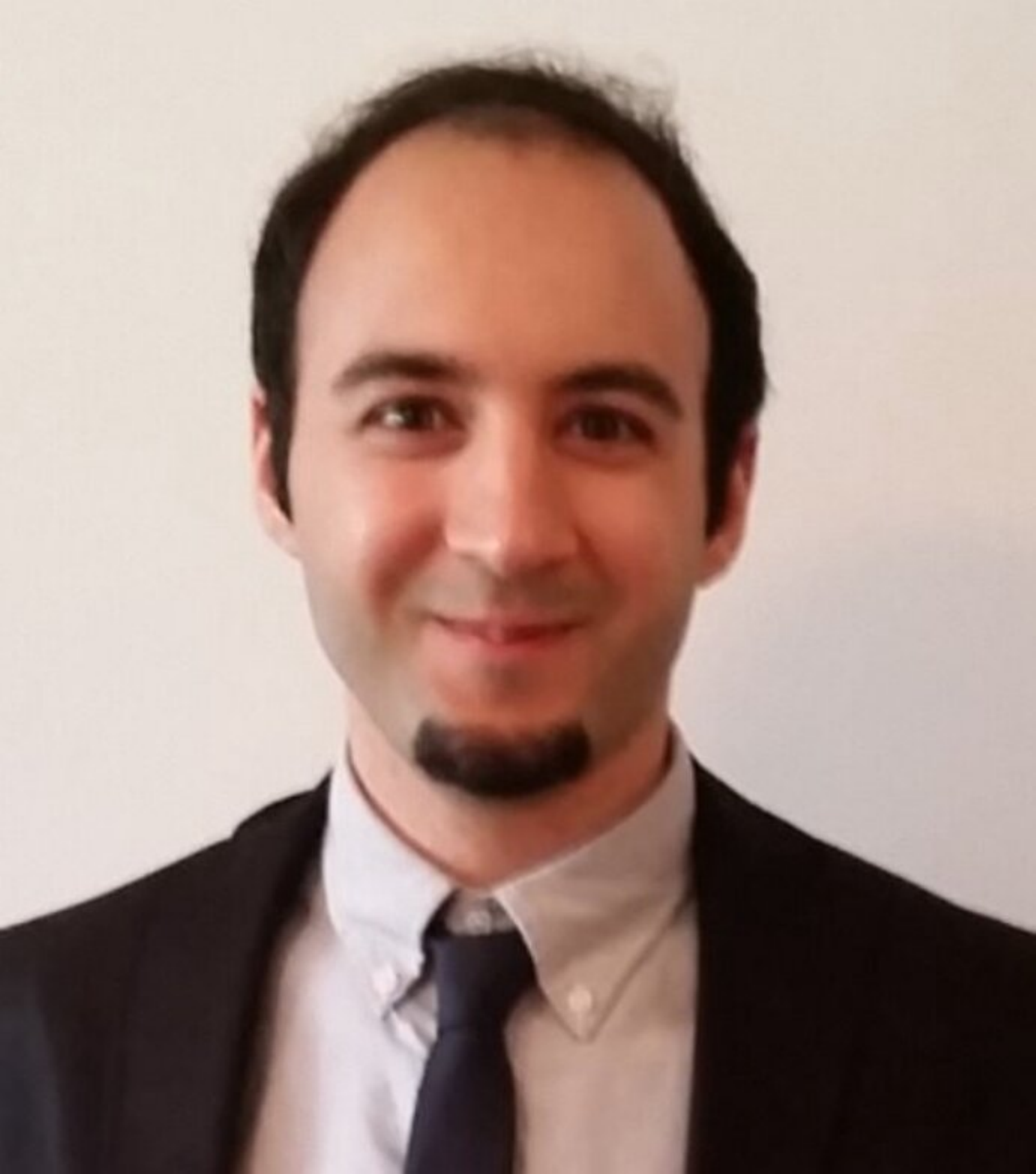}}]{Shervin Minaee} is a machine learning lead in the computer vision team at Snapchat, Inc. 
He received his PhD in Electrical Engineering and Computer Science from New York University, in 2018. 
His research interests include computer vision, image segmentation, biometric recognition, and applied deep learning.
He has published more than 40 papers and patents during his PhD.
He previously worked as a research scientist at Samsung Research, AT\&T Labs, Huawei Labs, and as a data scientist at Expedia group.
He has been a reviewer for more than 20 computer vision related journals from IEEE, ACM, Elsevier, and Springer. He has won several awards, including the best research presentation at Samsung Research America in 2017 and the Verizon Open Innovation Challenge Award in 2016.
\end{IEEEbiography}

\vskip -5pt plus -1fil

\begin{IEEEbiography}[{\includegraphics[width=0.95in,height=1.25in,clip,keepaspectratio]{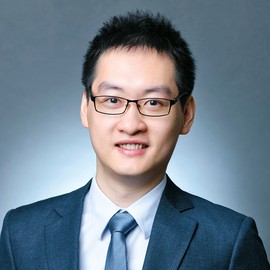}}]{Ping Luo} is an Assistant Professor in the department of computer science, The University of Hong Kong (HKU). He received his PhD degree in 2014 from Information Engineering, the Chinese University of Hong Kong (CUHK), supervised by Prof. Xiaoou Tang and Prof. Xiaogang Wang. He was a Postdoctoral Fellow in CUHK
from 2014 to 2016. He joined SenseTime Research as a Principal Research Scientist from 2017 to 2018. His research interests are machine learning and computer vision. He has published 70+ peer-reviewed articles in top-tier conferences and journals
such as TPAMI, IJCV, ICML, ICLR, CVPR, and NIPS. His work has high impact with 7,000 citations according to Google Scholar. He has won a number of competitions and awards such as the first runner up
in 2014 ImageNet ILSVRC Challenge, the first place in 2017 DAVIS Challenge on Video Object Segmentation, Gold medal in 2017 Youtube 8M Video Classification Challenge, the first place in 2018 Drivable Area
Segmentation Challenge for Autonomous Driving, 2011 HK PhD Fellow Award, and 2013 Microsoft Research Fellow Award (ten PhDs in Asia)
\end{IEEEbiography}

\vskip -5pt plus -1fil

\begin{IEEEbiography}[{\includegraphics[width=0.95in,height=1.25in,clip,keepaspectratio]{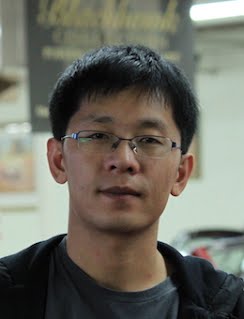}}]{Zhe Lin} is Senior Principal Scientist in Creative Intelligence Lab, Adobe Research. He received his Ph.D. degree in Electrical and Computer Engineering from University of Maryland at College Park in May 2009. Prior to that, he obtained his M.S. degree in Electrical Engineering and Computer Science from Korea Advanced Institute of Science and Technology in August 2004, and B.Eng. degree in Automation from University of Science and Technology of China. He has been a member of Adobe Research since May 2009. His research interests include computer vision, image processing, machine learning, deep learning, artificial intelligence. He has served as a reviewer for many computer vision conferences and journals since 2009, and recently served as an Area Chair for WACV 2018, CVPR 2019, ICCV 2019, CVPR 2020, ECCV 2020, ACM Multimedia 2020.
\end{IEEEbiography}

\vskip -5pt plus -1fil

\begin{IEEEbiography}[{\includegraphics[width=0.95in,height=1.25in,clip,keepaspectratio]{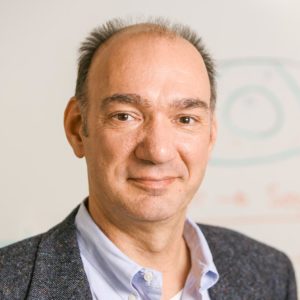}}]{Kevin W. Bowyer}  (Fellow, IEEE) received the Ph.D. degree in computer science from Duke University, Durham, NC, USA.
\\He is the Schubmehl-Prein Family Professor of computer science and engineering with the University of Notre Dame, Notre Dame, IN, USA.
\\Prof. Bowyer received the Technical Achievement Award from the IEEE Computer Society, with the citation “for pioneering contributions to the science and engineering of biometrics.” He served as the Editor-in-Chief for the IEEE TRANSACTIONS ON PATTERN ANALYSIS AND MACHINE INTELLIGENCE. He is currently serving as the Editor-in-Chief for the IEEE TRANSACTIONS ON BIOMETRICS, BEHAVIOR AND IDENTITY SCIENCE. In 2019, he was elected as a fellow of the American Association for the Advancement of Science (AAAS).
\end{IEEEbiography}

\end{document}